\documentclass[sigconf]{acmart}

\usepackage{xcolor}
\usepackage{tikz}
\usetikzlibrary{calc}
\usetikzlibrary{bayesnet}
\usetikzlibrary{arrows}
\usetikzlibrary{automata}
\usetikzlibrary{calc}
\usepackage{color}
\usepackage[export]{adjustbox}
\usepackage{subcaption}
\usepackage{mathtools}

\usepackage{enumitem}
\usepackage{booktabs}
\usepackage{array}
\usepackage{todonotes}

\AtBeginDocument{%
  \providecommand\BibTeX{{%
    \normalfont B\kern-0.5em{\scshape i\kern-0.25em b}\kern-0.8em\TeX}}}

\copyrightyear{2020} 
\acmYear{2020} 
\setcopyright{acmlicensed}\acmConference[ICAIF '20]{ACM International Conference on AI in Finance}{October 15--16, 2020}{New York, NY, USA}
\acmBooktitle{ACM International Conference on AI in Finance (ICAIF '20), October 15--16, 2020, New York, NY, USA}
\acmPrice{15.00}
\acmDOI{10.1145/3383455.3422520}
\acmISBN{978-1-4503-7584-9/20/10}

\begin{document}

\title{Financial Table Extraction in Image Documents}

\author{William Watson}
\affiliation{%
  \institution{S\&P Global}
}
\email{william.watson@spglobal.com}

\author{Bo Liu}
\authornote{Work done while affiliated with S\&P Global}
\affiliation{%
  \institution{NVIDIA}
 }
\email{boli@nvidia.com}




\begin{abstract}
   Table extraction has long been a
   pervasive problem in financial services. This is more challenging in the
   \textit{image} domain, 
   where content is locked behind cumbersome pixel format.
   Luckily, advances in deep learning for image segmentation, OCR, and 
   sequence modeling provides the necessary
   heavy lifting to achieve impressive results.
   This paper presents an end-to-end pipeline for \textit{identifying}, \textit{extracting} and 
   \textit{transcribing} tabular content in \textit{image} documents, 
   while retaining the original spatial relations with high fidelity.
\end{abstract}



\maketitle

\section{Introduction}
\begin{figure}
  \begin{center}
    \includegraphics[width=0.41\columnwidth, valign=c]{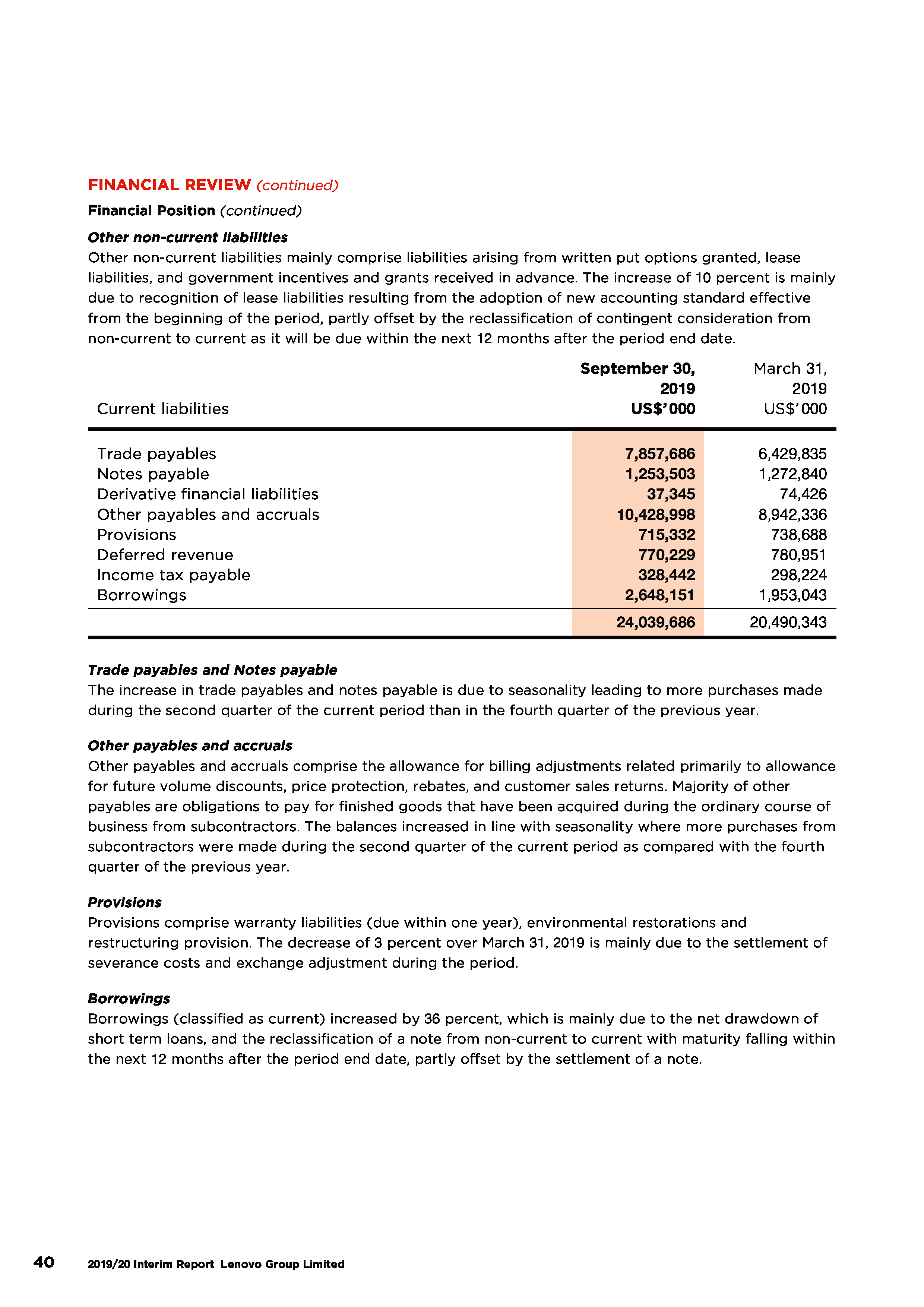}
    \includegraphics[width=0.41\columnwidth, valign=c]{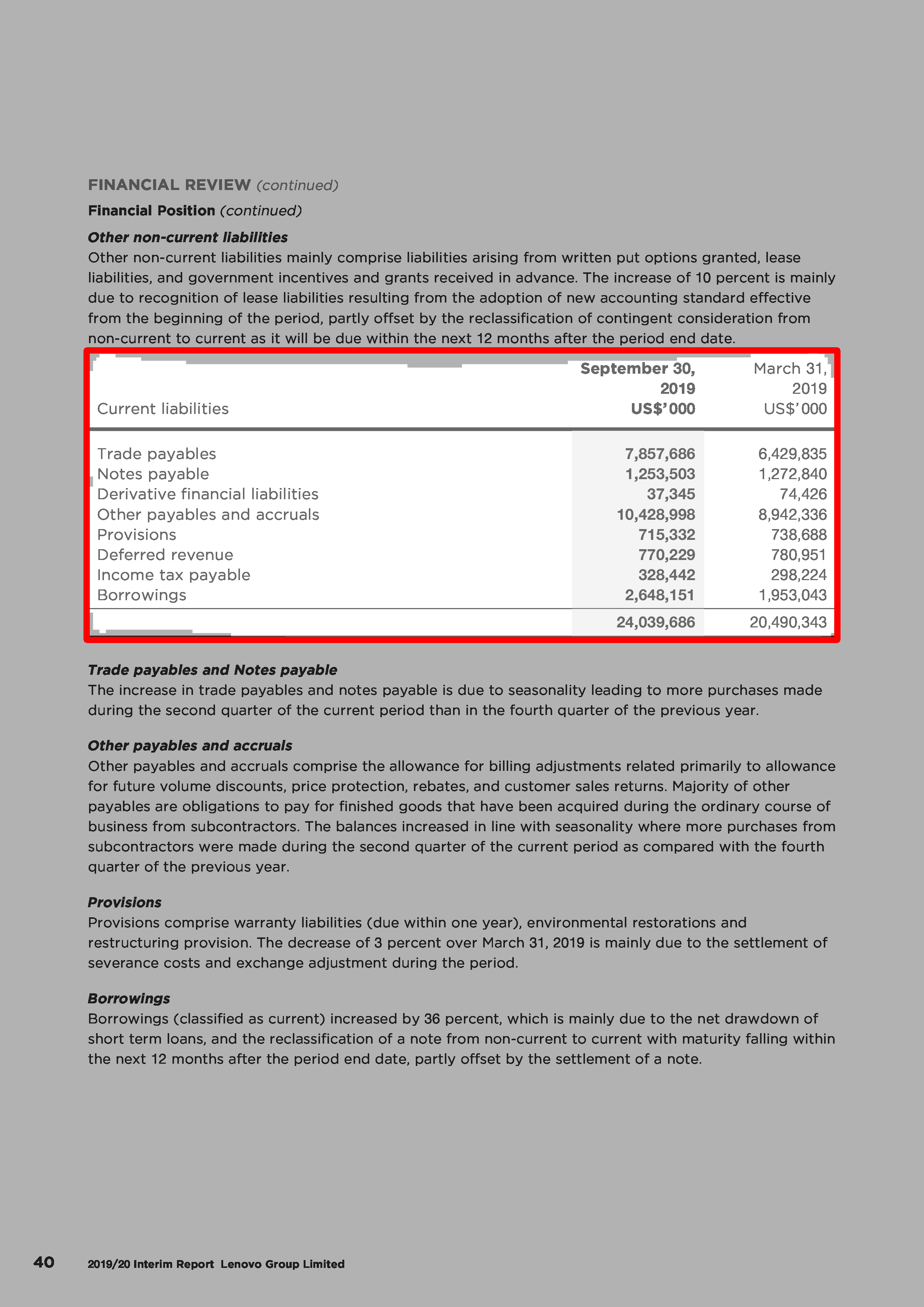}

    \includegraphics[width=0.94\columnwidth, valign=c]{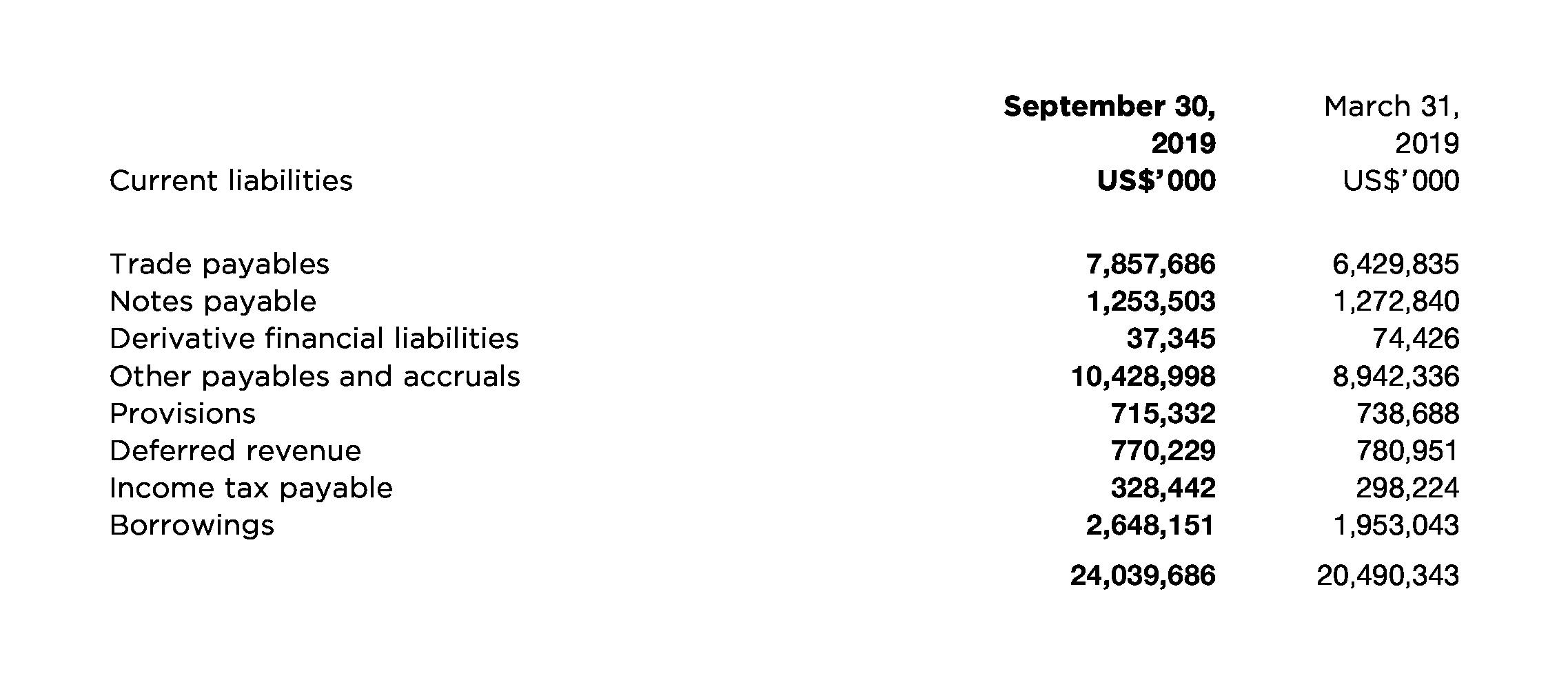}

    \resizebox{0.84\columnwidth}{!}{
    \begin{tabular}{lrr}
      \toprule
                                        &  September 30, &   March 31, \\
                                        &           2019 &        2019 \\
                    Current liabilities &       US\$’000 &     US\$’000 \\
                         Trade payables &      7,857,686 &   6,429,835 \\
                          Notes payable &      1,253,503 &   1,272,840 \\
      Derivative financial liabilities &         37,345 &      74,426 \\
            Other payables and accruals &     10,428,998 &   8,942,336 \\
                             Provisions &        715,332 &     738,688 \\
                      Deferred revenue &        770,229 &     780,951 \\
                     Income tax payable &        328,442 &     298,224 \\
                             Borrowings &      2,648,151 &   1,953,043 \\
                                        &     24,039,686 &  20,490,343 \\
      \bottomrule
    \end{tabular}
    }
    \end{center}
    \caption{The three steps of tabular extraction from image docs. Top: table identification. Middle: content recovery. Bottom: alignment into tabular format.} \label{fig:pipeline}
    \vspace{-3mm}
  \end{figure}

Table extraction has long been a persistent problem in automating data
collection in the financial service sector. Documents vary in style and content,
each being derived from a unique source. This paper concerns itself with the
\textit{image} domain, where the only machine-readable data is \textit{pixelated}.
These financial documents do not have a consistent format
and often exhibit high variance in layouts across pages and files.
Therefore, financial image documents pose several unique challenges:
\begin{enumerate}
    \item How can we find a table?
    \item How can we recover the content of a table?
    \item How can we align the content into a tabular format?
\end{enumerate}

Our proposed pipeline effectively segments the problem into these
three sub-tasks. 
Luckily, deep learning offers several techniques to tackle these sub-problems,
and when built in conjunction with
classical data structures and algorithms, can be shown to create
an effective pipeline for table extraction.
But one must ask, \textit{why do we need a system like this?} 
First, images are \textit{pixelated}, a cumbersome format for text extraction and processing.
For these formats, there is no way to copy and paste the content. 
Techniques such as optical character recognition (OCR) or human transcription are required to digitize content
into a friendlier format, to ascribe meaning to a set of colored pixels. 
Hence, this paper proposes an image based pipeline to
transcribe tabular
information locked in \textit{image} format into a variety of \textit{structured} outputs (CSV, Dataframes, or \LaTeX).

\section{Relevant Background}
\label{sec:background}


Table detection is not a new problem. 
There is an extensive body of prior literature that have used various
approaches in detecting tables in images.
Kieninger et al.~\cite{t_rec_1,t_rec_2} used a system called T-Recs to take 
word boundaries and cluster them into a segmentation graph to detect tables. 
However, it failed in the presence of multi-column layouts. 
Wang et al.~\cite{wang} used the distance between consecutive words 
as a heuristic in determining table entity candidates, but was tied to
a specific layout template.
Hu et al.~\cite{hu} proposed an approach involving single columned images.
Shafait et al.~\cite{table_detection_shafait} looked at heterogeneous 
documents, and their system was eventually integrated into Tesseract.
Tupaj et al.~\cite{tupaj} searches the image for sequences of 
table-like lines based on header keywords. 
Harit et al.~\cite{harit} used unique table start and trailer patterns,
but fails when the patterns are unique. 
Gatos et al.~\cite{gatos} uses the area of intersection between 
horizontal and vertical lines to reconstruct the intersectional pairs.
However, this system relies on the visual cues provided by strict, defining table borders. 
Hidden Markov Models were proposed by Costa e Silva~\cite{silva_hmm},
and Kasar~\cite{kasar} locates tables through column and row separators 
with Support Vector Machines. 
Other methods were explored by Jahan et al.~\cite{jahan} with word 
spacing and line height thresholds for table localization.
Hybrid approaches attempt to discover table candidate regions, as in 
Anh et al.~\cite{anh}.
Finally, deep learning methods were presented in 
Hao et al.~\cite{hao_region_proposal} regional proposal network
using CNNs, and Gilani et al.~\cite{table_detection_rcnn} adapted the Faster R-CNN
architecture to segment table regions in a given image.
We propose a pipeline for \textit{financial image} documents that 
\textit{identifies} table boundaries, \textit{extracts} text content,
and \textit{aligns} table cells using spatial, textual, and sequential features.





\section{Table Detection}
\label{sec:detection}

Table detection converts PDF files into images and 
detects table boundaries for each page as a list of coordinates $(x1,x2,y1,y2)$.
Even though this is a detection task, we adopt a semantic segmentation approach 
for two reasons. 
First, object detection models are designed to detect multiple categories of objects in the same image, as the case for benchmark detection tasks such as COCO~\cite{lin2014microsoft}. 
But here we only care about one type of object, \textit{tables}. 
Semantic segmentation solves the problem of classifying each pixel into categories, i.e. segmenting the image into different regions. 
Therefore our problem can be framed as a segmentation task with two categories, the \textit{table region} and \textit{other}. 
Second, object detection models (especially two-stage ones like Faster R-CNN~\cite{ren2015faster} and Cascade R-CNN~\cite{cai2017cascade}) are much heavier than the semantic segmentation model U-Net \cite{unet}, in terms of both training and inference time.
Since a segmentation model's output masks can be of any shape, post-processing is performed to convert them into proper bounding boxes.

\subsection{Models}


The first step is to convert arbitrary length PDFs into single page images. We used ImageMagick's \texttt{convert} tool \footnote{https://imagemagick.org/script/convert.php} at 300 dpi
grayscale with no compression loss (PNG). 
We chose 300 dpi in order to ensure that images during the OCR phase (\S\ref{sec:ocr}) remain
at an acceptable quality for text extraction.

\subsubsection{Classification Model}
\label{sec:page_class}

In financial documents, not all pages have tables. Depending on the corpus, the proportion of \textit{empty} pages (i.e., no tables) is about half. 
Although segmentation models can deal with empty pages by predicting zero tables after post-processing, a classification model works better for the task of identifying empty pages. 
Therefore we use a binary table classification model to filter pages without tables before feeding the remaining pages into the segmentation model.
Annotating pages as a binary task is much cheaper than annotating table locations for segmentation, so we can easily create a large training set covering a variety of non-table pages, such as those with different types of pictures and graphs.
Inference is twice as fast with the classification model over the segmentation model.
Our model uses the Inception-ResNet
v2~\cite{res_inception} convolution base with input size $384\times288\times1$ plus a dense layer of dimension 256. Light augmentation was applied in training: random zoom within $\pm10\%$ and random horizontal/vertical shift within $\pm5\%$ to imitate poor quality scans.

\begin{figure}
    \centering
    \includegraphics[width=0.60\linewidth]{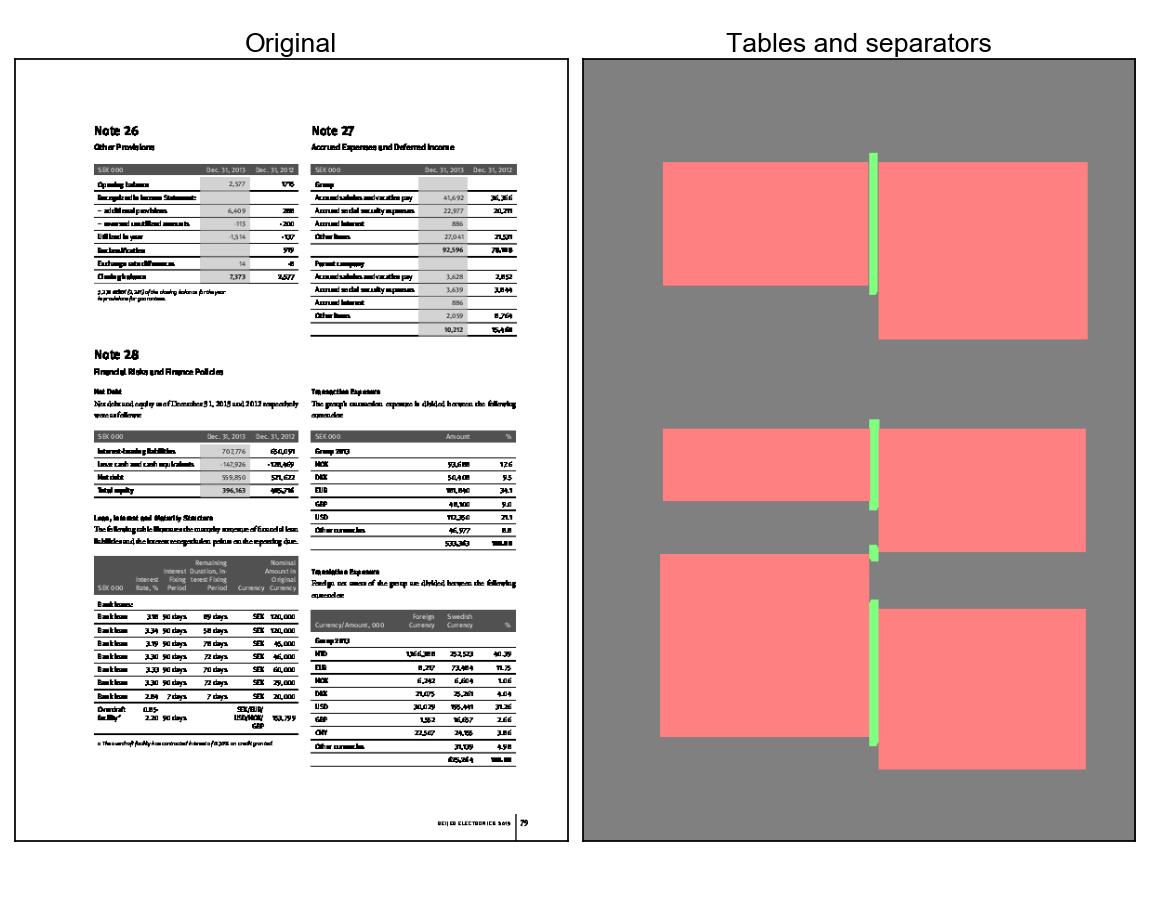}
    \caption{The \textit{separator} trick: when tables are close to each other, generate a third class, separator (light green), from annotated table boxes (red).}
    \label{fig:separator_example}
    \vspace{-4mm}
\end{figure}

\subsubsection{Segmentation Model}
\label{sec:table_seg}
Our pipeline adopts U-Net~\cite{unet}, which was originally developed for biomedical image segmentation and widely used in various domain tasks such as autonomous driving~\cite{siam2018comparative} and satellite imagery segmentation~\cite{demir2018deepglobe,iglovikov2017satellite}.
In particular, we used a U-Net architecture with DenseNet-169~\cite{huang2017densely} backbone and softmax activation, similar to the winning solution\footnote{\url{https://github.com/selimsef/dsb2018_topcoders}} of 2018 Data Science Bowl~\cite{caicedo2019nucleus}. The input size is $384 \times 288 \times 3$. The loss function is a combination of categorical cross entropy (60\%) and dice coefficient loss (20\% each for the table and separator channels)~\cite{sudre2017generalised}.
For augmentation, we applied 2\textdegree~random rotation and 0.01 shear intensity to imitate low-quality scans.
A limitation of applying semantic segmentation to solve object detection problems is that the model does not differentiate instances of the same object type---semantically, they are the same object. Therefore, the model would struggle when multiple objects of the same type are close together, or even touching. A common trick used in nucleus segmentation for biomedical microscopy image analysis is to add a \textit{boundary} class, 
resulting in a 3-class semantic segmentation task: \textit{background}, \textit{nucleus boundary}, and \textit{inside nucleus} \cite{cui2019deep,caicedo2019nucleus}. 
Inspired by this, we add \textit{separators} between closely located tables, making our problem 3-class: \textit{table}, \textit{separator}, and \textit{other}. 
We do not do full table boundaries as in nucleus segmentation because in a typical nucleus microscopy image, there are dozens of nuclei squeezed together taking up most of the area, making the full boundary necessary. 
But in our case, all tables are rectangular and most are not in close proximity to others. 
Therefore we only need to add separators when necessary, as in Figure \ref{fig:separator_example}. After annotation, we used scikit-image's \cite{scikit-image} \textit{dilation} and \textit{watershed} functions to automatically generate the separators. 
Separators appeared in 13.4\% of our annotated set.


\subsubsection{Post-processing}
\hspace{0pt}

\textit{Thresholding}. The raw output of U-Net is the probability of each pixel belonging to a table. A threshold can convert these into a binary mask: the smaller the threshold the larger the mask and vice versa. We tuned the threshold on our validation set such that output masks have maximal \textit{intersection over union} (IOU) with ground truth bounding boxes. 
The optimal threshold is 0.7 at 0.84 IOU.

\textit{Separating tables}. 
We use scikit-image to separate the raw masks into different connected regions. 
Each connected mask is a table candidate. We remove masks that are smaller than 1\% of total image area. This is based on the observation that such masks are generally noise as all tables are larger than 1\% of the whole page.

\textit{Rectanglization}. Most individual raw masks are near rectangular. We need to convert them into proper bounding boxes. We do this by taking the minimal enclosing rectangle for each raw mask. The resulting bounding box is always larger than the raw mask. But since we used tight bounding boxes when annotating tables, the results turned out to be good for most cases (see \S\ref{sec:detection_results}).

\subsection{Dataset and Labeling}

\begin{figure*}
  \centering
  \vspace{-3mm}
    \begin{minipage}{.5\textwidth}
      \centering
      \captionof{figure}{Full Success: Border Lines}
      \label{fig:S1}
      \includegraphics[height=2.59cm]{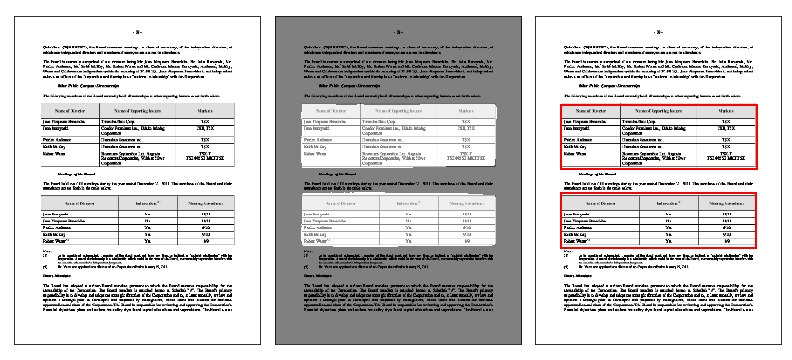}
    \end{minipage}%
    \begin{minipage}{.5\textwidth}
      \centering
      \captionof{figure}{Full Success: No Border Lines}
      \label{fig:S2}
      \includegraphics[height=2.59cm]{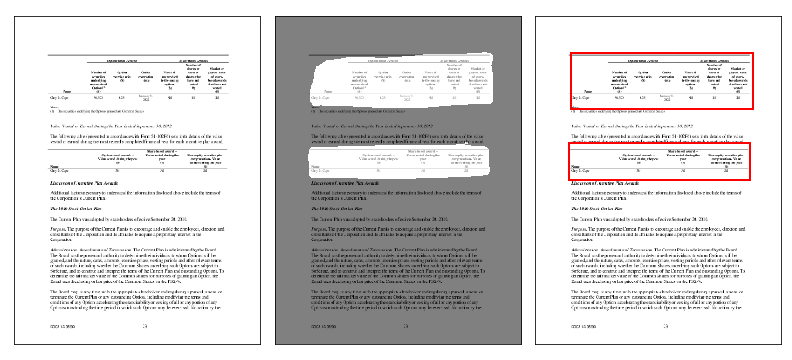}
    \end{minipage}%
    
    \vspace{-3mm}
    \begin{minipage}{.5\textwidth}
      \centering
      \captionof{figure}{Full Success: Inline Table}
      \label{fig:S3}
      \includegraphics[height=2.59cm]{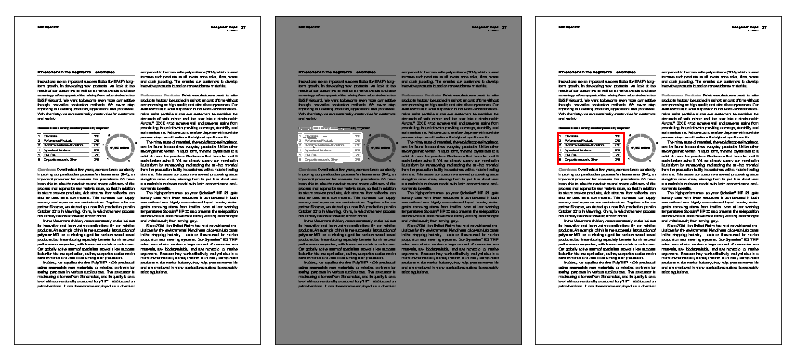}
    \end{minipage}%
    \begin{minipage}{.5\textwidth}
      \centering
      \captionof{figure}{Full Success: Stacked Tables}
      \label{fig:S4}
      \includegraphics[height=2.59cm]{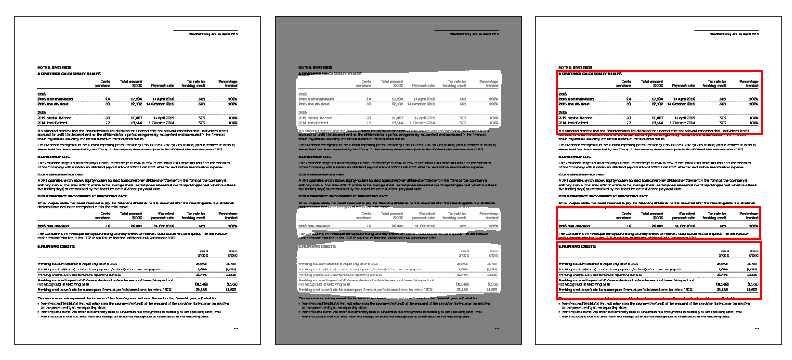}
    \end{minipage}%
    
    \vspace{-3mm}
    \begin{minipage}{.5\textwidth}
      \centering
      \captionof{figure}{Partial Success: Multiple Tables}
      \label{fig:P1}
      \includegraphics[height=2.59cm]{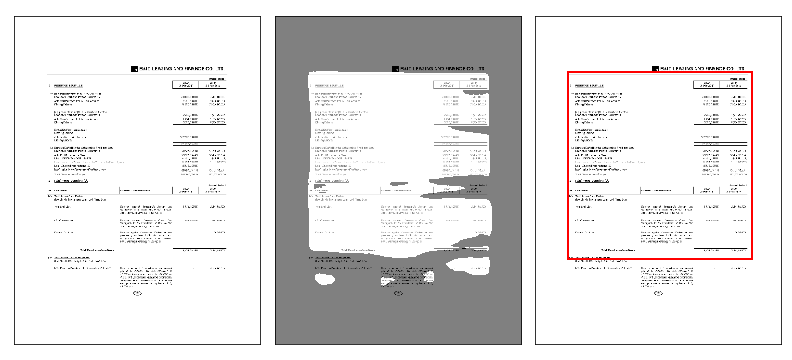}
    \end{minipage}%
    \begin{minipage}{.5\textwidth}
      \centering
      \captionof{figure}{Partial Success: Multiple Tables}
      \label{fig:P2}
      \includegraphics[height=2.59cm]{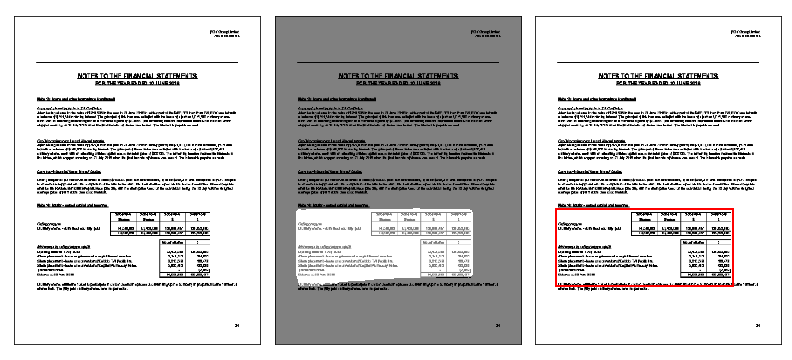}
    \end{minipage}%
    
    \vspace{-3mm}
    \begin{minipage}{.5\textwidth}
      \centering
      \captionof{figure}{Partial Success: Nearby Non-Tabular Text}
      \label{fig:P3}
      \includegraphics[height=2.59cm]{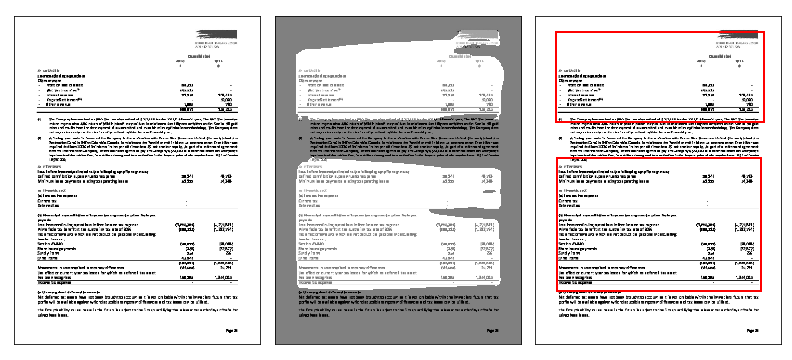}
    \end{minipage}%
    \begin{minipage}{.5\textwidth}
      \centering
      \captionof{figure}{Partial Success: Incomplete Table}
      \label{fig:P4}
      \includegraphics[height=2.59cm]{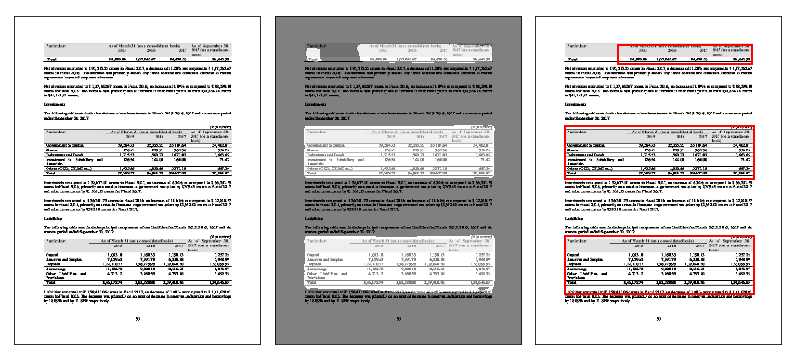}
    \end{minipage}%
    \vspace{-1mm}
\end{figure*}

We sampled 10,000 pages from a collection of 9,985 financial documents, including annual reports, prospectus documents and shareholder meeting notes published by international companies in various sectors such as commodities, banking, and technology. 
The dataset was curated for its diversity of content and reporting style.
Pages were labeled as either \textit{having tables} (4,593) or \textit{no tables} (5,407). 
8,000 are used as the training set and 2,000 for validation.
Separately, for segmentation, we labeled the bounding boxes for all tables on 909 random pages with tables using the annotation tool LabelMe \cite{labelme2016}. There were 1,660 tables in total, or 1.8 tables per page. The max number of tables on a page was 10. 
The U-Net model is trained on a smaller set comprised of 749 pages with 160 for validation, for two reasons. 
First, for classification, each annotated page only provides 1 bit of information (has a table or not), but a U-Net model can learn more information from a fully annotated segmentation page---each \textit{pixel} is 1 bit of information (this pixel belongs to a table or not).
Second, annotating bounding boxes is more time-consuming than annotating whether a page contains tables.
After development, we selected another 1,000 random pages to test the robustness of the pipeline. 
We did not annotate bounding boxes because IOU is \textit{not sufficient} to tell whether the detection is satisfactory on a given page, e.g. predicting extra blank surrounding area is more tolerable than missing the table header. Instead, we plotted the predicted bounding boxes on each page (figures~\ref{fig:S1}---\ref{fig:P4}) and judged if the detection was fully correct, partially correct, or incorrect. 

\subsection{Results}\label{sec:detection_results}

For the classification model, recall is more important than precision: false positives are acceptable since the downstream U-Net model, with post-processing, has the ability to predict no tables on a page; but false negatives mean a page with tables is thrown away before going further in the pipeline. We tuned the threshold on the validation set such that recall is 0.995 and precision is 0.871. 

We examined the pipeline's output on 1,000 test pages. On the test set, recall is 0.993 and precision is 0.910. Note that recall and precision are table level metrics. 
On the page-level, 83\% were completely correct, followed by 
9.5\% covering multiple tables, 5.9\% covering nearby text, and 
1.6\% representing other errors. 
Note that the latter three cases are still considered \textit{partially correct}.


Figures~\ref{fig:S1}---\ref{fig:S4} show some \textit{correct} cases. 
Figure~\ref{fig:S1} is the easiest: tables with borders on all sides, resulting in near rectangular masks. 
Tables in figure~\ref{fig:S2} either have few or no borders, resulting in non-rectangular masks. 
Nonetheless, the masks cover all the table's content and post-processing recovers the correct area. 
Figure~\ref{fig:S3} is an inline table,
and the segmentation has some holes in the output mask, which post-processing handles easily. 
Figure~\ref{fig:S4} are the cases where two tables are located next to each other with minimal space in between. 
The separator trick is crucial for such cases---the results were much worse before. 
Here, the segmentation model successfully separated the tables' masks even though the shapes are not perfect. 


Figures \ref{fig:P1}---\ref{fig:P4} shows \textit{partially correct} cases. 
The most common error is the accidental merger of multiple tables.
In figure \ref{fig:P1}, there are two tables with no space or text between them. 
Neither table has borders. 
The segmentation model predicted a merged blob which was post-processed into a single bounding box. 
In figure \ref{fig:P2}, there are two inline tables next to each other, 
and the model predicted a single bounding box covering both tables, plus additional text on the left. 
Another issue is when the predicted bounding box covers nearby text, as shown in figure \ref{fig:P3}. 
The top bounding box includes a line of text below and the top-right page header. 
Figure~\ref{fig:P4} is an example of a less common case, where for very wide but short tables, the segmentation model missed some columns if spacing between columns is large.
We want to emphasize that none of the partially correct examples are a complete failure. 
Most examples can be handled by downstream steps to different degrees, but we were conservative in evaluation and did not count them as completely correct, to give a fair assessment of the detection models.

\section{Table Extraction}
\label{sec:ocr}


\subsection{Tesseract OCR}
Given the coordinates of a segmented table, we can \textit{extract}
textual information alongside positional metadata from images with Tesseract OCR.
It is a vital component that bridges the 
gap between table detection and alignment.
Tesseract utilizes an LSTM-based engine to predict the character
for a given segment of pixels, built alongside line and word finding
algorithms~\cite{tesseract_main}. In addition, 
Tesseract provides practical tools for script, language, and orientation
detection, allowing us to extract content in other languages, or 
content that is rotated~\cite{tesseract_main,tesseract_orientation}. 
In addition, Tesseract provides
useful layout information for lines, blocks, and word boundaries, 
which is convenient in organizing text into tabular format~\cite{tesseract_layout}.

\subsection{Preprocessing Pipeline}
\label{sec:preprocess}

Given the boundaries of a segmented table, our preprocessing pipeline to enhance the
image before OCR consists of three major components: 
Otsu Thresholding~\cite{otsu}, Automatic Orientation Correction,
and Morphological Kernel Filtering.
Prior to these steps, we also crop and pad the region of interest to avoid
boundary issues.
Our implementation is built from well-known computer vision recipes using OpenCV~\cite{opencv}.

\begin{table*}
  \centering
    \caption{Additional Table Pipeline Examples}\label{table:edge_cases}
    \vspace{-3mm}
    \begin{tabular}{ c  c  c  c  c }
      \toprule
      {} & Original & Segmentation Mask & OCR Input & Final Alignment \\
      \cmidrule(r){2-2}\cmidrule(lr){3-3}\cmidrule(lr){4-4}\cmidrule(l){5-5}
      A
      &
      \begin{minipage}{.12\textwidth} 
        \includegraphics[width=\linewidth]{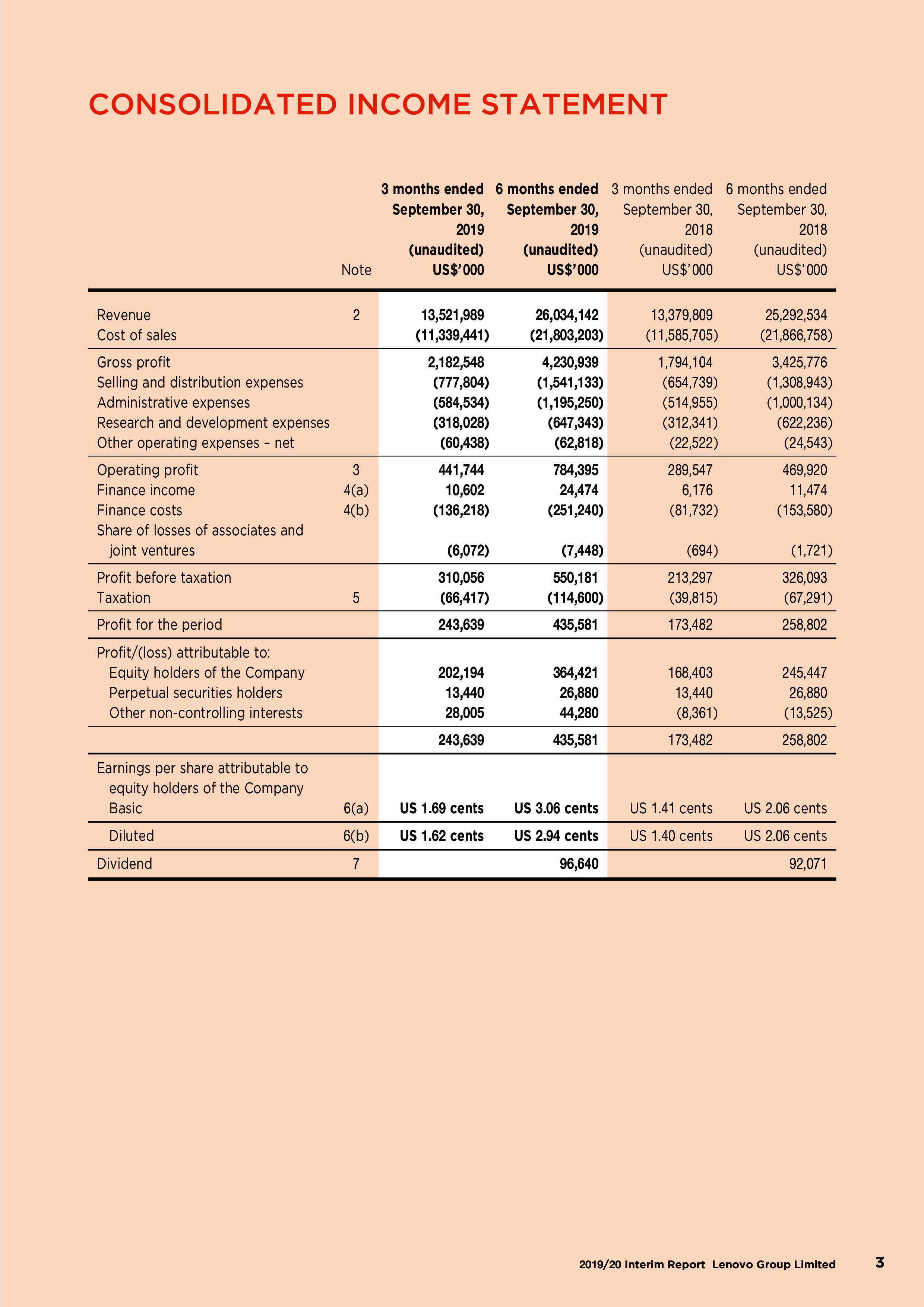}
      \end{minipage}
      &
      \begin{minipage}{.12\textwidth} 
        \includegraphics[width=\linewidth]{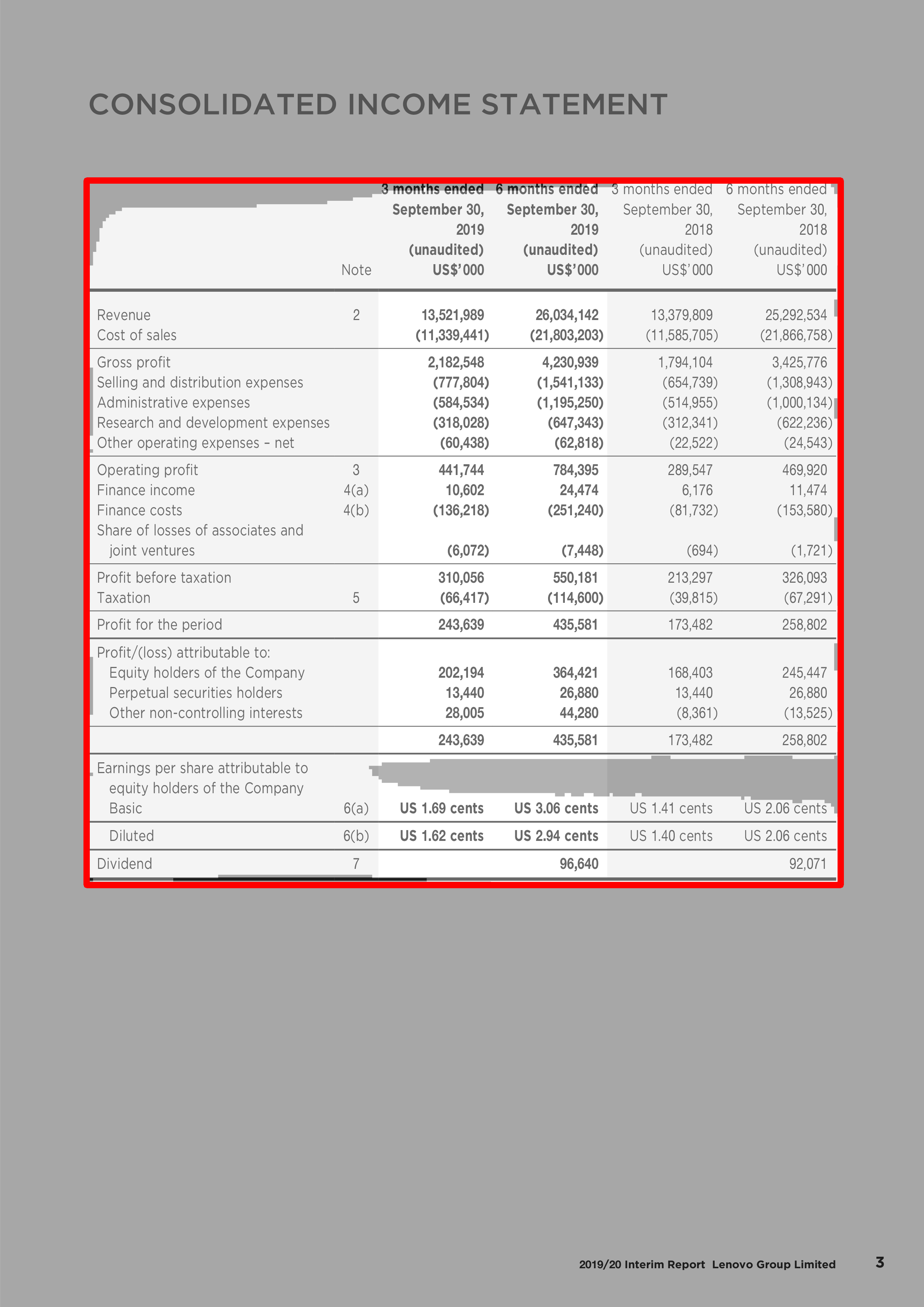}
      \end{minipage}
      &
      \begin{minipage}{.15\textwidth}
        \includegraphics[width=\linewidth]{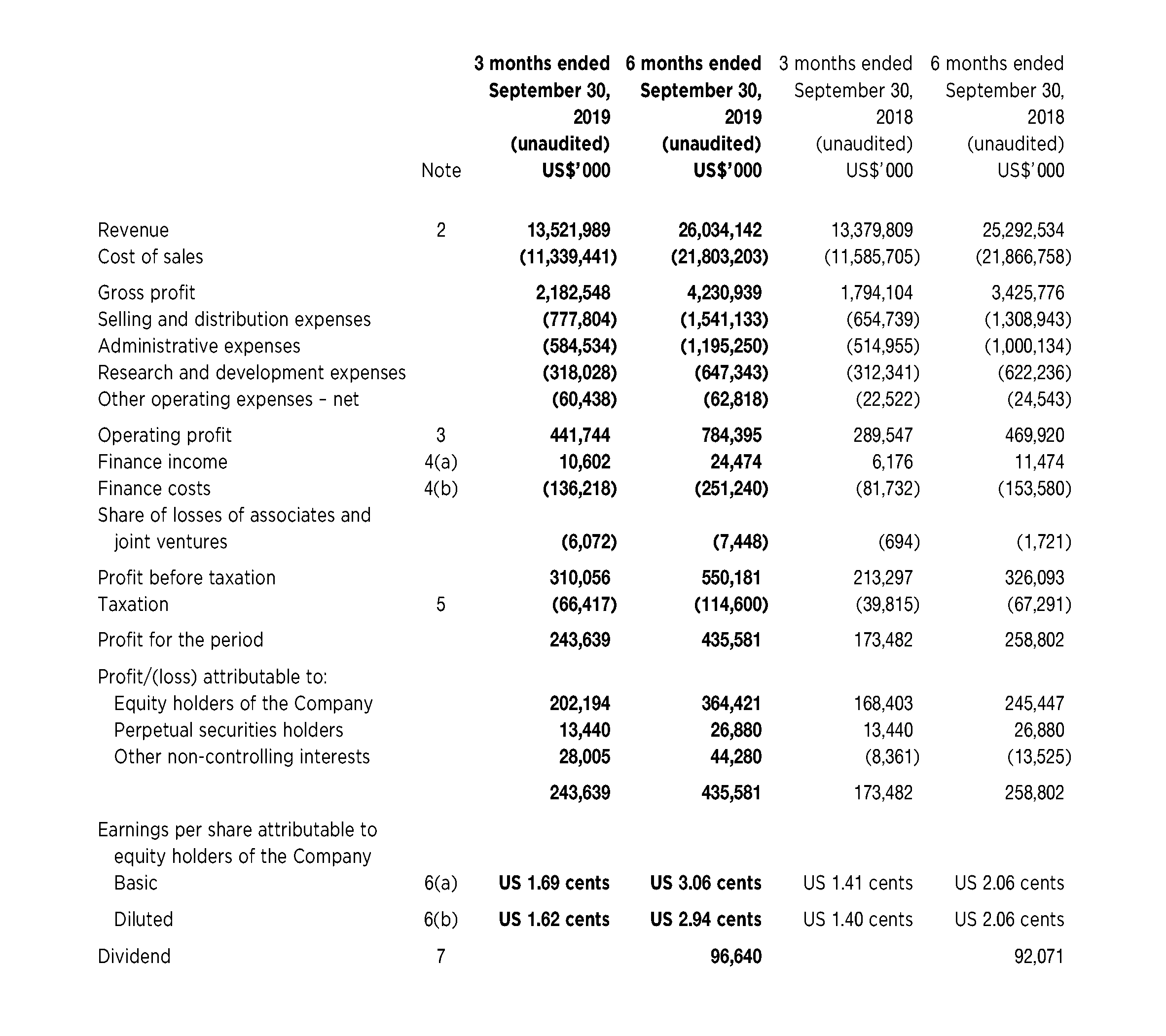}
      \end{minipage}
      &
      \resizebox{.15\textwidth}{!}{
        \begin{tabular}{lcrrrr}
\toprule
                                    &       &  3 months ended &  6 months ended &  3 months ended &  6 months ended \\
                                    &       &   September 30, &                 &   September 30, &   September 30, \\
                                    &       &            2019 &            2019 &            2018 &            2018 \\
                                    &       &     (unaudited) &     (unaudited) &     (unaudited) &     (unaudited) \\
                                    &  Note &         US\$’000 &         US\$’000 &                 &                 \\
                            Revenue &     2 &      13,521,989 &      26,034,142 &      13,379,809 &      25,292,534 \\
                      Cost of sales &       &    (11,339,441) &    (21,803,203) &    (11,585,705) &    (21,866,758) \\
                       Gross profit &       &       2,182,548 &       4,230,939 &       1,794,104 &       3,425,776 \\
  Selling and distribution expenses &       &       (777,804) &     (1,541,133) &       (654,739) &     (1,308,943) \\
            Administrative expenses &       &       (584,534) &     (1,195,250) &       (514,955) &     (1,000,134) \\
  Research and development expenses &       &       (318,028) &       (647,343) &       (312,341) &       (622,236) \\
     Other operating expenses - net &       &        (60,438) &        (62,818) &        (22,522) &        (24,543) \\
                   Operating profit &     3 &         441,744 &         784,395 &         289,547 &         469,920 \\
                     Finance income &  4(a) &          10,602 &          24,474 &           6,176 &          11,474 \\
                      Finance costs &  4(b) &       (136,218) &       (251,240) &        (81,732) &       (153,580) \\
  Share of losses of associates and &       &                 &                 &                 &                 \\
                     joint ventures &       &         (6,072) &         (7,448) &           (694) &         (1,721) \\
             Profit before taxation &       &         310,056 &         550,181 &         213,297 &         326,093 \\
                           Taxation &     5 &        (66,417) &       (114,600) &        (39,815) &        (67,291) \\
              Profit for the period &       &         243,639 &         435,581 &         173,482 &         258,802 \\
     Profit/(loss) attributable to: &       &                 &                 &                 &                 \\
      Equity holders of the Company &       &         202,194 &         364,421 &         168,403 &         245,447 \\
       Perpetual securities holders &       &          13,440 &          26,880 &          13,440 &          26,880 \\
    Other non-controlling interests &       &          28,005 &          44,280 &         (8,361) &        (13,525) \\
                                    &       &         243,639 &         435,581 &         173,482 &         258,802 \\
 Earnings per share attributable to &       &                 &                 &                 &                 \\
      equity holders of the Company &       &                 &                 &                 &                 \\
                              Basic &  6(a) &   US 1.69 cents &   US 3.06 cents &   US 1.41 cents &   US 2.06 cents \\
                            Diluted &  6(b) &   US 1.62 cents &   US 2.94 cents &   US 1.40 cents &   US 2.06 cents \\
                           Dividend &     7 &                 &          96,640 &                 &          92,071 \\
\bottomrule
\end{tabular}

      }
      \\ \midrule 
      B
      &
      \begin{minipage}{.12\textwidth} 
        \includegraphics[width=\linewidth]{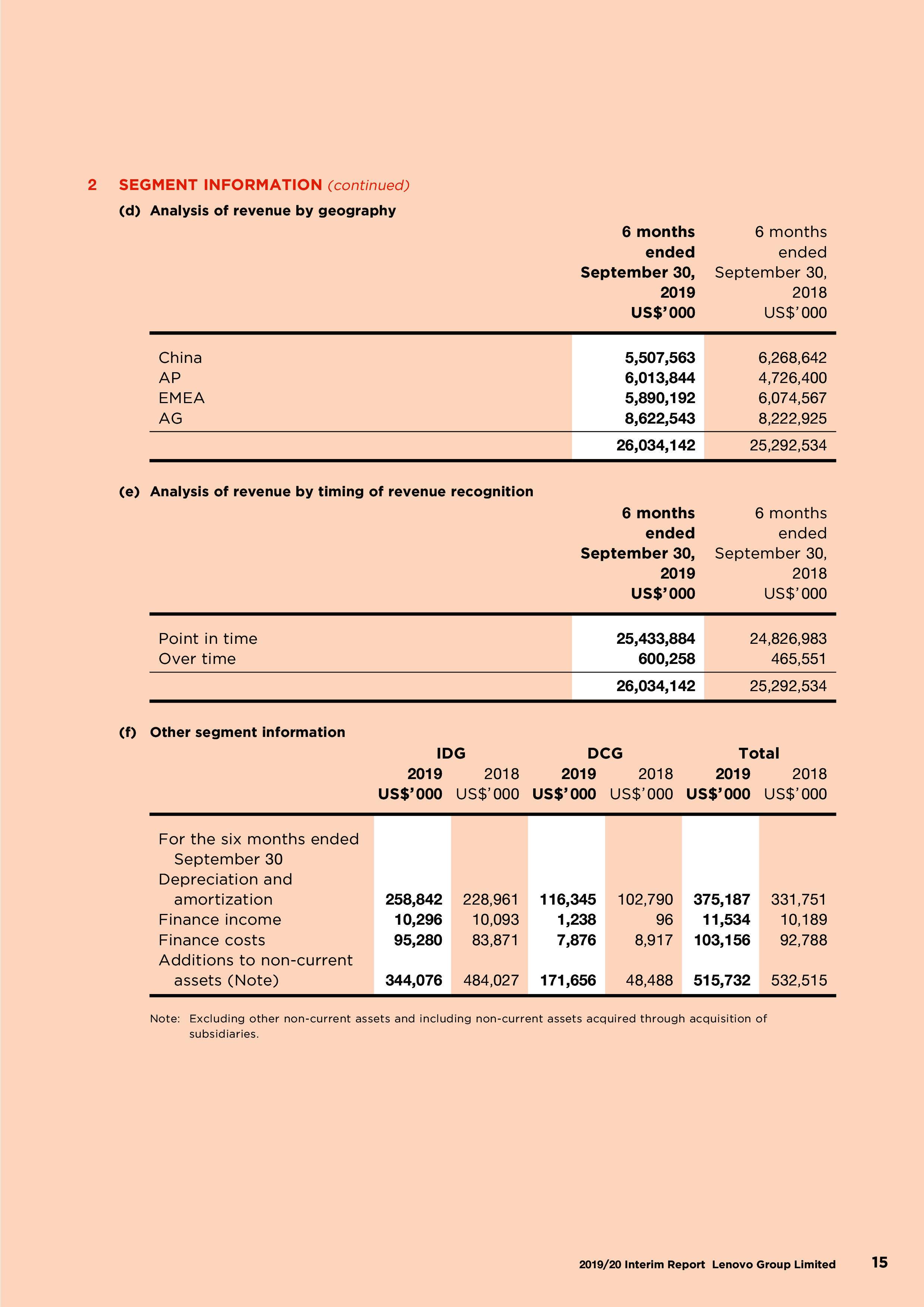}
      \end{minipage}
      &
      \begin{minipage}{.12\textwidth} 
        \includegraphics[width=\linewidth]{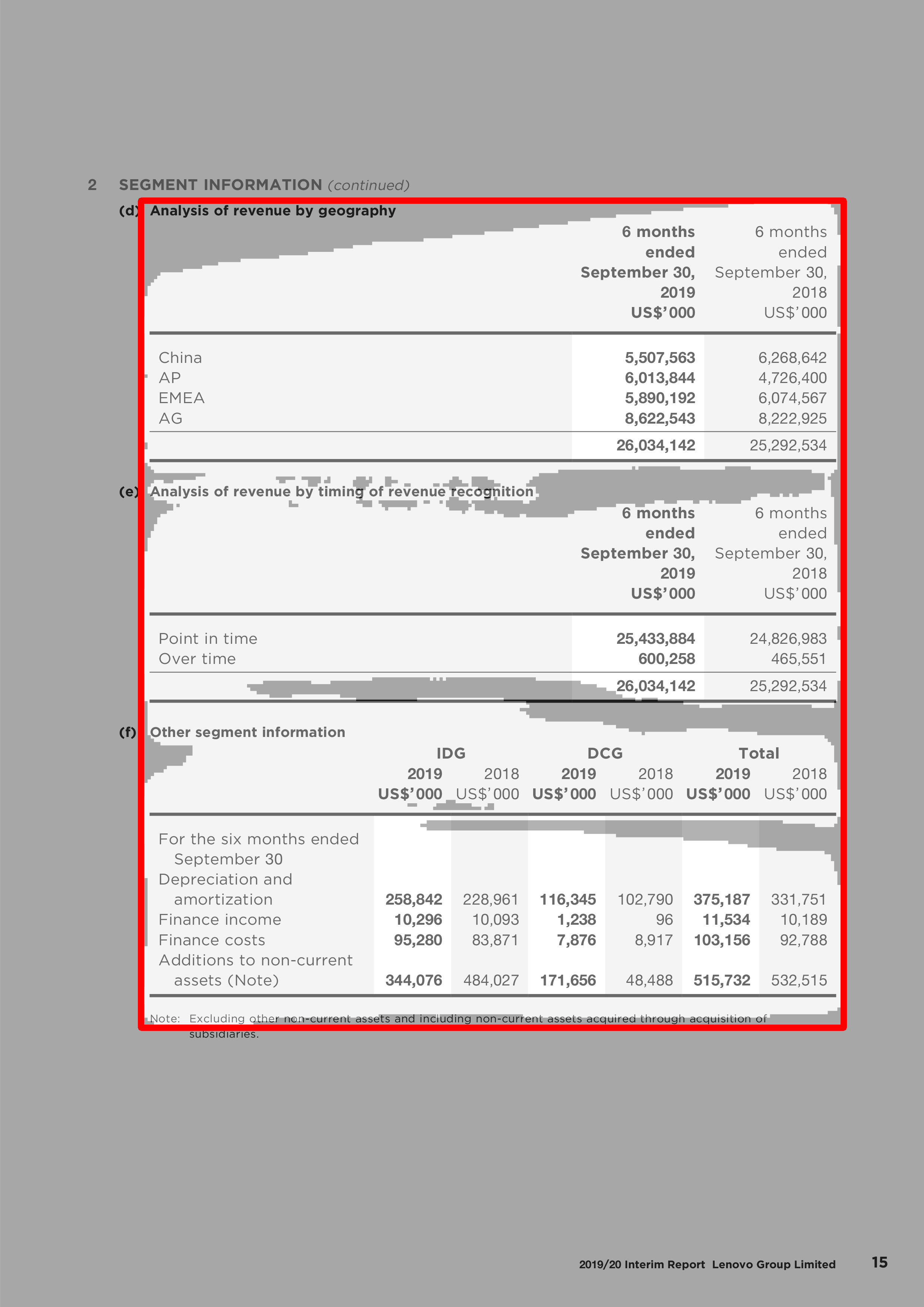}
      \end{minipage}
      &
      \begin{minipage}{.14\textwidth}
        \includegraphics[width=\linewidth]{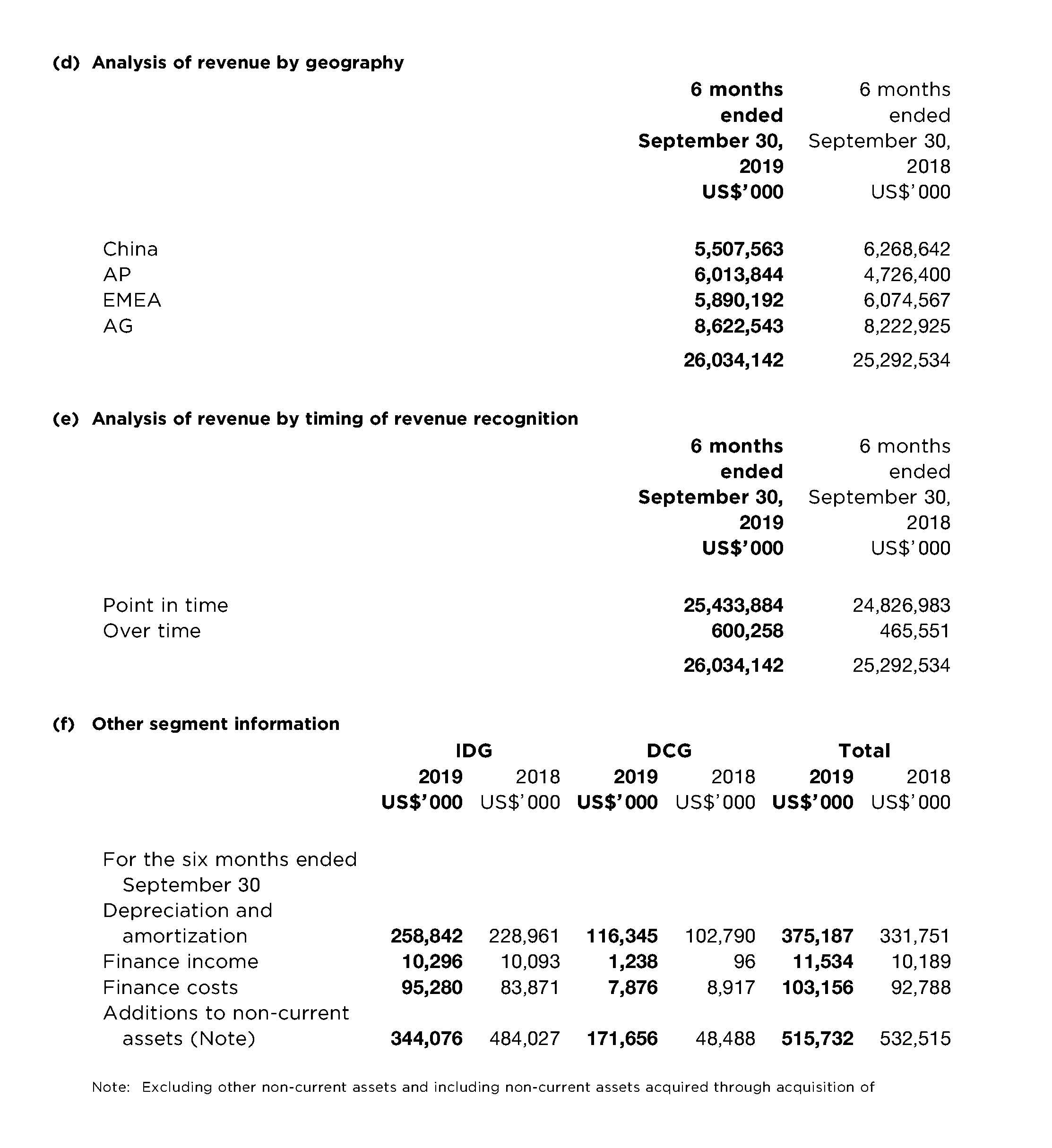}
      \end{minipage}
      &
      \resizebox{.14\textwidth}{!}{
        \begin{tabular}{lrrcrrrr}
\toprule
 (d) &                   Analysis of revenue by geography &          &      &           &          &                &                  \\
     &                                                    &          &      &           &          &       6 months &         6 months \\
     &                                                    &          &      &           &          &          ended &            ended \\
     &                                                    &          &      &           &          &  September 30, &    September 30, \\
     &                                                    &          &      &           &          &           2019 &             2018 \\
     &                                                    &          &      &           &          &        US\$’000 &         US\$’ 000 \\
     &                                              China &          &      &           &          &      5,507,563 &        6,268,642 \\
     &                                                 AP &          &      &           &          &      6,013,844 &        4,726,400 \\
     &                                               EMEA &          &      &           &          &      5,890,192 &        6,074,567 \\
     &                                                 AG &          &      &           &          &      8,622,543 &        8,222,925 \\
     &                                                    &          &      &           &          &     26,034,142 &       25,292,534 \\
 (e) &  Analysis of revenue by timing of revenue recog... &          &      &           &          &                &                  \\
     &                                                    &          &      &           &          &       6 months &         6 months \\
     &                                                    &          &      &           &          &          ended &            ended \\
     &                                                    &          &      &           &          &  September 30, &    September 30, \\
     &                                                    &          &      &           &          &           2019 &             2018 \\
     &                                                    &          &      &           &          &        US\$’000 &         US\$’ 000 \\
     &                                      Point in time &          &      &           &          &     25,433,884 &       24,826,983 \\
     &                                          Over time &          &      &           &          &        600,258 &          465,551 \\
     &                                                    &          &      &           &          &     26,034,142 &       25,292,534 \\
 (f) &                          Other segment information &          &      &           &          &                &                  \\
     &                                                    &          &  IDG &           &          &            DCG &            Total \\
     &                                                    &     2019 &      &      2018 &     2019 &           2018 &        2019 2018 \\
     &                                                    &  US\$’000 &      &  US\$’'000 &  US\$’000 &        US\$’000 &  US\$’000 US\$’000 \\
     &                           For the six months ended &          &      &           &          &                &                  \\
     &                                       September 30 &          &      &           &          &                &                  \\
     &                                   Depreciation and &          &      &           &          &                &                  \\
     &                                       amortization &  258,842 &      &   228,961 &  116,345 &        102,790 &  375,187 331,751 \\
     &                                     Finance income &   10,296 &      &    10,093 &    1,238 &             96 &    11,534 10,189 \\
     &                                      Finance costs &   95,280 &      &    83,871 &    7,876 &          8,917 &   103,156 92,788 \\
     &                           Additions to non-current &          &      &           &          &                &                  \\
     &                                      assets (Note) &  344,076 &      &   484,027 &  171,656 &         48,488 &  515,732 532,515 \\
     &  Note: Excluding other non-current assets and i... &          &      &           &          &                &                  \\
\bottomrule
\end{tabular}

      }
      \\ \midrule 
      C
      &
      \begin{minipage}{.12\textwidth} 
        \includegraphics[width=\linewidth]{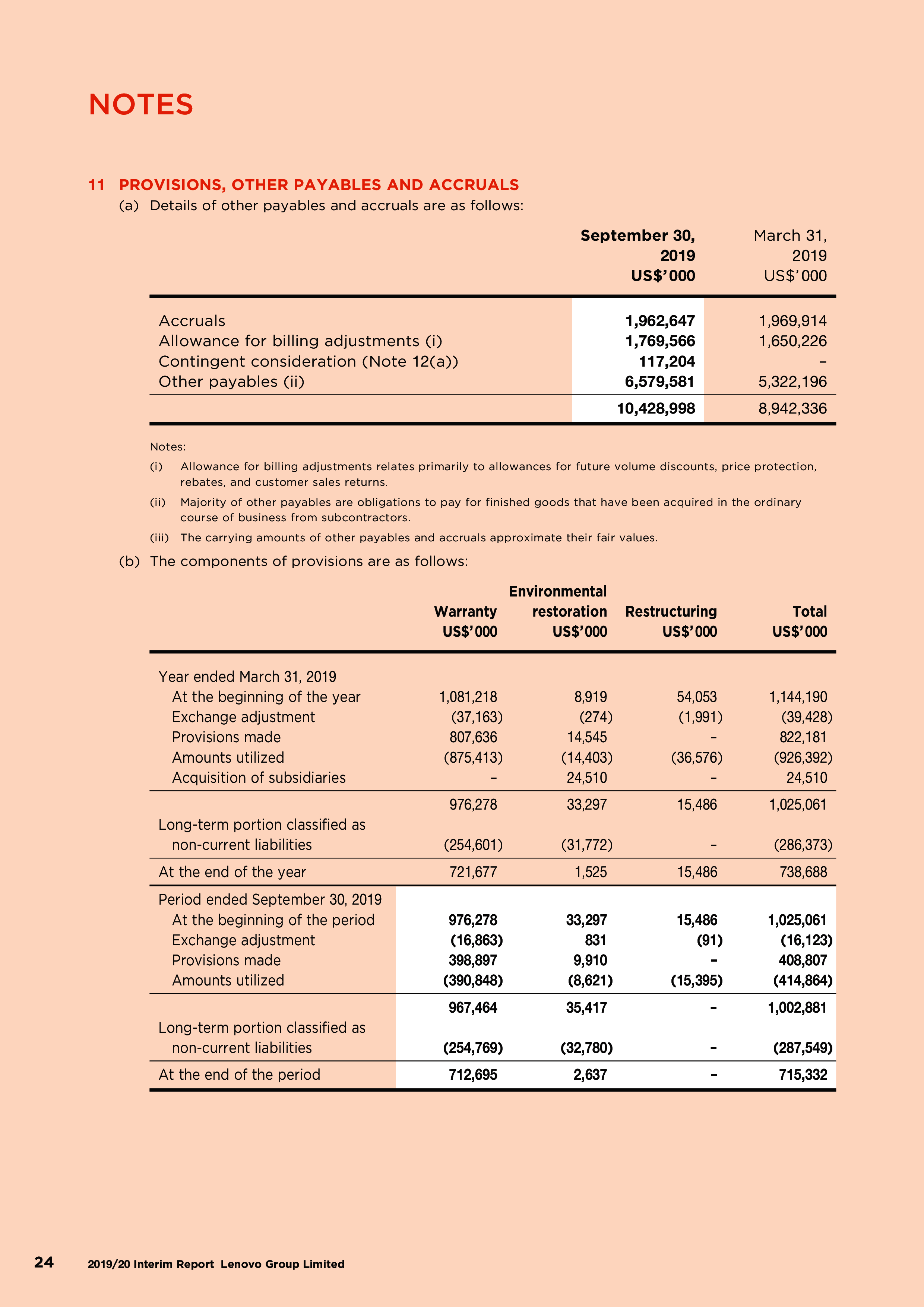}
      \end{minipage}
      &
      \begin{minipage}{.12\textwidth} 
        \includegraphics[width=\linewidth]{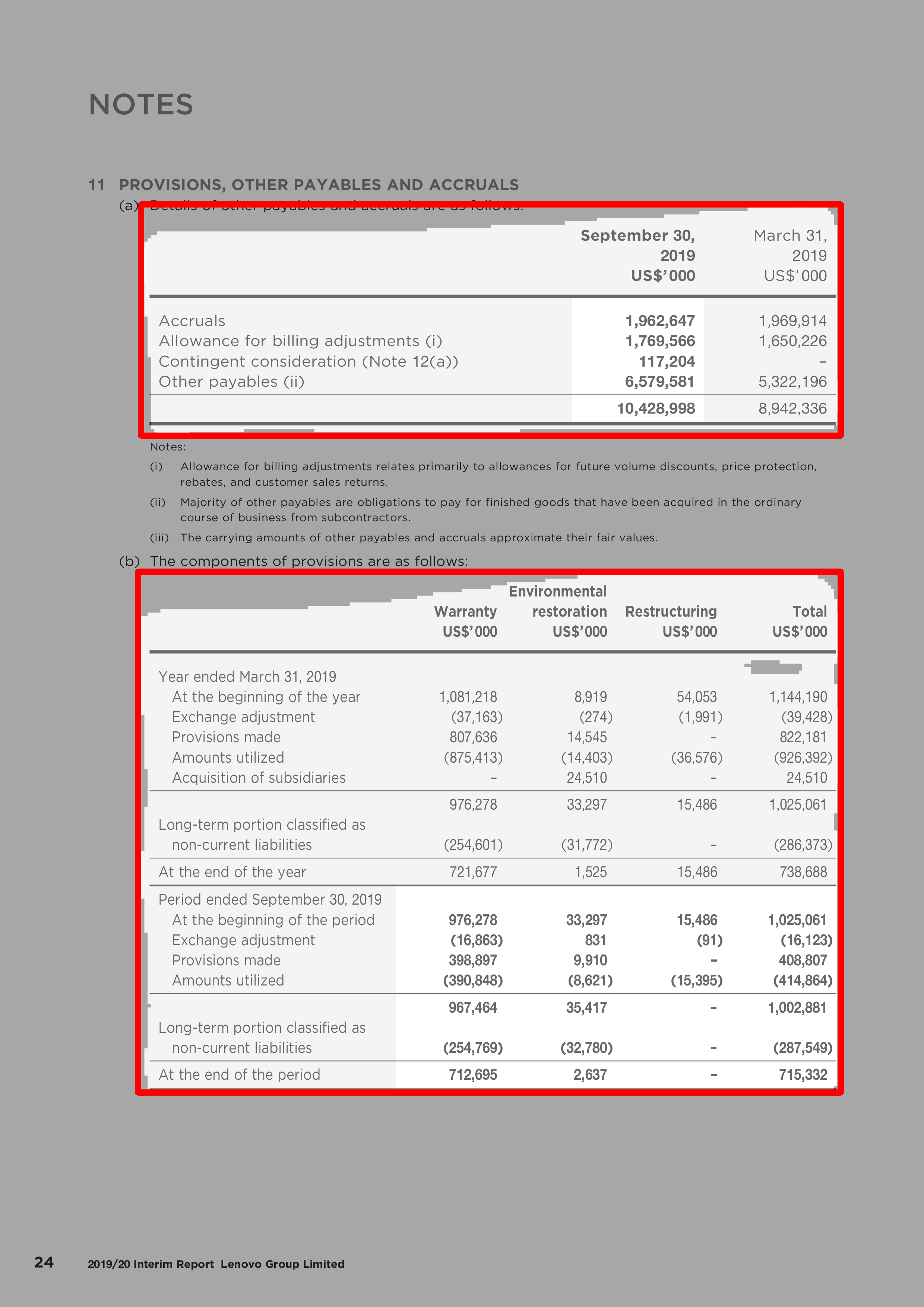}
      \end{minipage}
      &
      \begin{minipage}{.13\textwidth}
        \includegraphics[width=\linewidth]{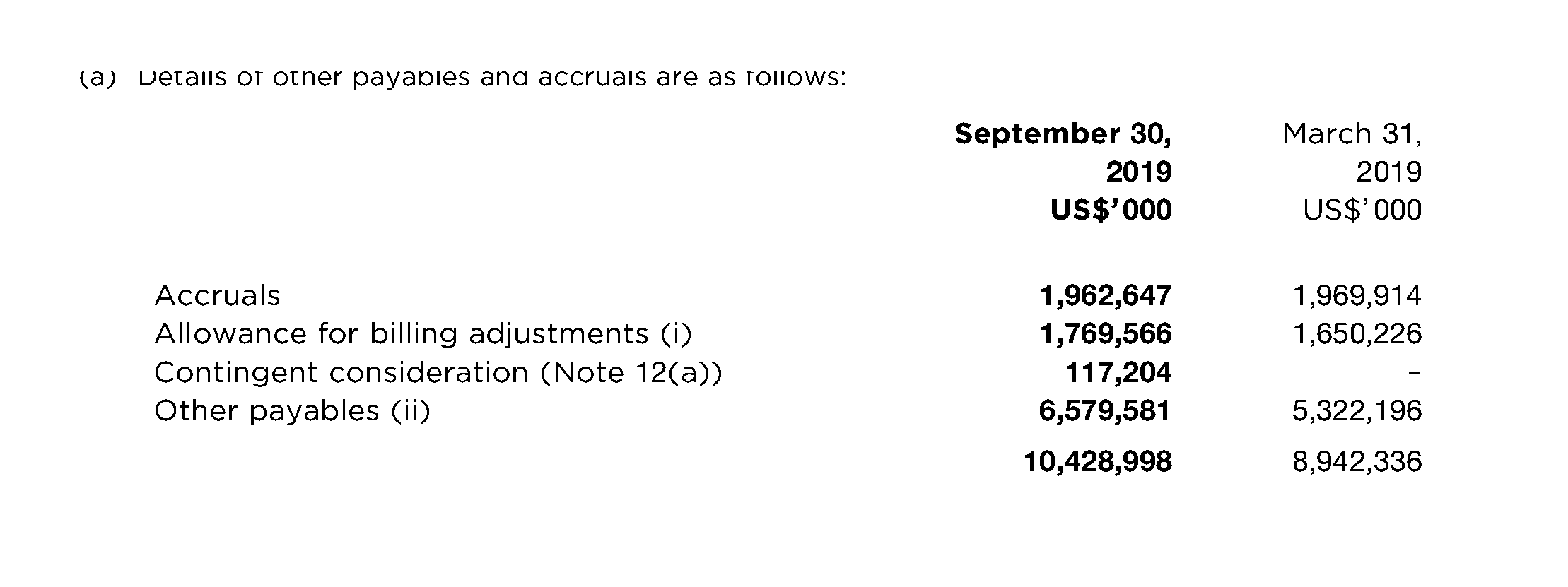}

        \vspace{0.3cm}

        \includegraphics[width=\linewidth]{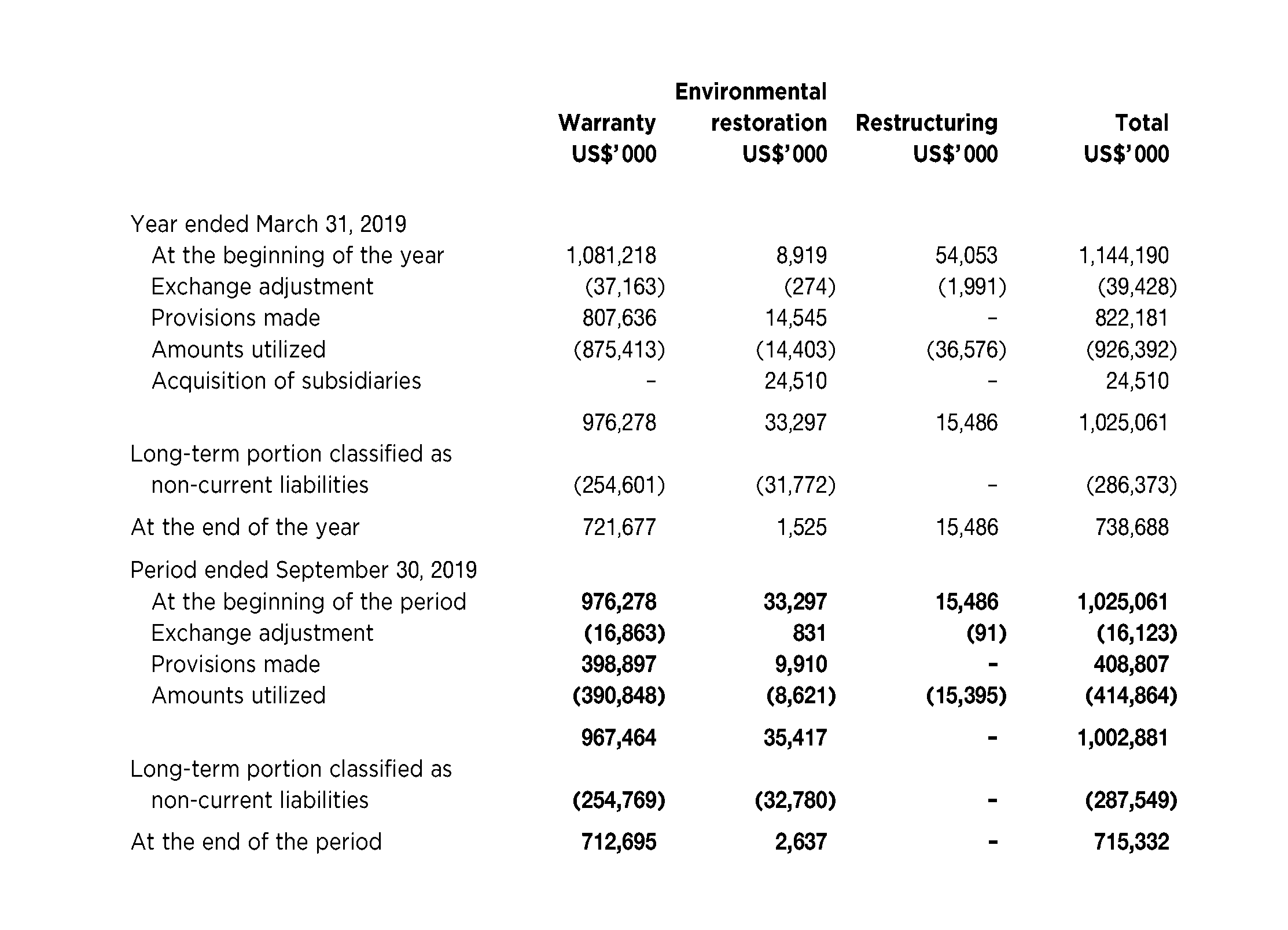}
      \end{minipage}
      &
      \begin{minipage}{.13\textwidth}

      \resizebox{\linewidth}{!}{
        \begin{tabular}{llrrr}
\toprule
 \{a) &  DVtails of otner payables and accruals are as... &                &              \\
     &                                                    &  September 30, &  March 31,    \\
     &                                                    &           2019 &       2019    \\
     &                                                    &        US\$’000 &   US\$’ 000  \\
     &                                           Accruals &      1,962,647 &  1,969,914    \\
     &              Allowance for billing adjustments Ci) &      1,769,566 &  1,650,226    \\
     &              Contingent consideration (Note 12(a)) &        117,204 &          -    \\
     &                                Other payables (ii) &      6,579,581 &  5,322,196    \\
     &                                                    &     10,428,998 &  8,942,336    \\
\bottomrule
\end{tabular}

      }

      \vspace{0.35cm}

      \resizebox{\linewidth}{!}{
        \begin{tabular}{lrrrr}
\toprule
                                 &            &   Environmental &                &            \\
                                 &   Warranty &     restoration &  Restructuring &      Total \\
                                 &            &         Us\$’000 &        Us\$’000 &    US\$’000 \\
       Year ended March 31, 2019 &            &                 &                &            \\
    At the beginning of the year &  1,081,218 &           8,919 &         54,053 &  1,144,190 \\
             Exchange adjustment &   (37,163) &           (274) &        (1,991) &   (39,428) \\
                 Provisions made &    807,636 &          14,545 &              - &    822,181 \\
                Amounts utilized &  (875,413) &        (14,403) &       (36,576) &  (926,392) \\
     Acquisition of subsidiaries &         -  &         24,510 &              - &     24,510 \\
                                 &    976,278 &          33,297 &         15,486 &  1,025,061 \\
 Long-term portion classified as &            &                 &                &            \\
         non-current liabilities &  (254,601) &        (31,772) &              - &  (286,373) \\
          At the end of the year &    721,677 &           1,525 &         15,486 &    738,688 \\
 Period ended September 30, 2019 &            &                 &                &            \\
  At the beginning of the period &    976,278 &          33,297 &         15,486 &  1,025,061 \\
             Exchange adjustment &   (16,863) &             831 &           (91) &   (16,123) \\
                 Provisions made &    398,897 &           9,910 &              - &    408,807 \\
                Amounts utilized &  (390,848) &         (8,621) &       (15,395) &  (414,864) \\
                                 &    967,464 &          35,417 &              - &  1,002,881 \\
 Long-term portion classified as &            &                 &                &            \\
         non-current liabilities &  (254,769) &        (32,780) &              - &  (287,549) \\
        At the end of the period &    712,695 &           2,637 &              - &    715,332 \\
\bottomrule
\end{tabular}

      }
      \end{minipage}
      \\ 
      \bottomrule
    \end{tabular}
    \vspace{-1mm}
\end{table*}

\subsubsection{Automatic Orientation Correction}
\label{sec:rot}

We use Tesseract's automatic orientation detection algorithm~\cite{tesseract_orientation}
to determine the angle of the text in the image. This is necessary because some
tables are in landscape mode, but are not properly rotated in the document itself.
Therefore, some images are rotated at $90$ or $270$ degree angles. 
We do not rotate if the detection returns $0$ or $180$, as it has been 
our observation that $180$ is given in error and 
assume that no table is completely flipped.
It is important to note that we also preserve the dimensions and scope of the rotated 
table by adjusting the rotation matrix with a translation term.




\subsubsection{Morphological Kernel Filtering}
\label{sec:filter}

OCR systems are sensitive to noise, and in particular table borders that can
accidentally register as a character or word. To mitigate this problem, we use a
filtering algorithm based on \textit{structuring elements} and
\textit{morphological operators} to remove lines from a table.
A structuring element, or kernel, is a preset pattern represented
by a binary grid. For our experiments, the kernel size was set to $50\times i$ for the horizontal
direction, and to $i\times 50$ for the vertical direction, with $i\in\{1,2\}$.
We set the main dimension to $50$
in order to capture long continuous streaks and ignore smaller
character strokes, for both the vertical and horizontal directions.
The basic operators for morphological transforms are \textit{erosion} and
\textit{dilation}. For \textit{erosion}, a binary image thins out, as the pixel
will remain a $1$ if and only if the window at said pixel exactly matches the
kernel. \textit{Dilation}, in contrast, will expand the object if any pixel in the
window matches an element in the kernel. \textit{Opening} is erosion followed by dilation.
The process of extracting lines in a single direction (horizontal or vertical)
is independent from the other. The algorithm simply applies the summed result
of an \textit{opening} operator.
Hence, a mask for each direction is created. These masks are added together,
\textit{dilated} to increase the line coverage, and slightly blurred using
a small $3 \times 3$ filter. 
Finally, we remove the lines through a bitwise AND
operation on the original image and mask, removing vertical and horizontal strokes from the table.

\subsection{Extracted Metadata}

Tesseract provides crucial metadata to manipulate the OCR results into a
structured table: 
the \textit{word} itself, 
the bounding box (\textit{left}, \textit{top}, \textit{width}, \textit{height}),
the confidence score,
and finally, sequential layout data (paragraph, line, and word identifiers).
We have found that the paragraph and line fields roughly correspond to horizontally
aligned text fields, and the word position is useful for a rough ordering
of a line's text. However, the key data is the bounding box and text information.
We drop low confidence OCR results to remove excess noise from the
alignment stage. This significantly reduces false triggers and catastrophic misalignments.


\section{Table Alignment}
\label{sec:alignment}

\subsection{Alignment Data Structure \& Algorithm}
\label{sec:uf}
We approach the concept of alignment as creating a pool of \textit{disjoint sets},
where each \textit{set} can correspond to a column. This
idea of \textit{unifying} individual text segments 
into a set is inspired by Knight's work
on unification~\cite{knight_unification}.
Luckily, the \textit{disjoint-set} or \textit{union-find} structure 
marries sub nodes into co-relating trees that represent 
a set efficiently~\cite{CLRS,uf_first}. With path compression and union by rank,
we ensure the height of these trees are shallow, and has been proven to provide
a \textit{search} and \textit{merge} time 
complexity of $O\left(\alpha\left(n\right)\right)$,
where $\alpha\left(n\right)$ is the inverse Ackermann function
~\cite{uf_bounds_4,uf_bounds_1,uf_bounds_5,uf_bounds_3,uf_bounds_2}.
Since the inverse Ackermann function is bounded to be 
less than $4$ for practical sizes of $n$,
this implies the structure
performs operations approximately in constant time~\cite{CLRS}.

The algorithm uses the
\textit{Intersection Over Union} to trigger the merging of two trees 
based on the width overlap of their member cells. 
This avoids merging two non-aligned cells and its children. 
When we merge 2 cells, all other previously merged cells are now co-related
without any additional work. 
Using this lookup structure, we can sort our sets
and assemble them into a grid by iterating through the rows and placing each
column cell in its most plausible alignment. This is trivial since we now have
the rough dimensions of the grid we are making from unique sets created by the
\textit{union-find} structure.
The result is a \textit{structured} dataframe that
can be exported into various outputs such as CSV, Excel, or
\LaTeX~for further downstream applications or direct embedding.

We use our labels to develop a upper bound on the number of true columns and
cells for a given table. 
We calculate the symmetric mean absolute percentage error (SMAPE) per table to obtain 
error distributions for each model~\cite{armstrong1996,FLORES198693}.
SMAPE is bounded between $0$ and $200\%$, and 
under-forecasting is penalized more than over-forecasting. 
Effectively, this means that we prefer errors that will overestimate cells 
since underestimating leads to catastrophic mergers that are harder to correct.
We report the relevant quantiles and maximum error rates, alongside the percentage of perfect
alignments, i.e. zero error.
\subsection{Column Segmentation}

Table alignment relies on intelligently organizing disjoint segments of words into 
a collection of rows and columns. 
Rows are already organized by unique line identifiers from Tesseract.
For columns, we explored different model schemes to
provide rough segments that can be merged into larger cells to help in the alignment process.

We sampled 3,640 suspected tables from the detection pipeline, drawn from a collection 
of unstructured financial documents that 
included financial statements, 
compliance certificates, management presentations, and more. 
We hand labeled cell boundaries on tables with the lines removed
(\S\ref{sec:filter}).
Interestingly, 3,038 (83.5\%) tables would be considered \textit{valid} by
human labelers. The \textit{invalid} cases were graphs, pictures, or bullet lists misidentified as a table.
Filtering our dataset on these tables, we ran the 
OCR pipeline to receive word metadata. 
Together, these tables have 99,000 unique tokens, and 510,000 individual segments to align. To make training tractable, we employ a 2 step fallback tokenization scheme. First, we include a token if it has a cumulative frequency over 75.
Secondly, if the token fails to meet this threshold, 
we instead use its combined POS tag generated from spaCy\footnote{\url{https://spacy.io/api/annotation\#pos-tagging}}.
A combined POS tag is all the individual POS tags concatenated together, i.e. (45\%) becomes
\textit{PUNCT-NUM-SYM-PUNCT} (POS length 4). 
If a POS tag does not have a frequency over 20, or the combined length is over 7, 
we treat it
as an unknown token. This method yielded 881 unique token tags.
For spatial information, we define 4 main features: \textit{distance to next token},
\textit{distance to previous token}, \textit{start of a row}, and
\textit{end of a row}. Only 2 can be active at a time, 
since a token at the start of a row
does not have a distance to a previous token, and vice versa. 
Distances are normalized to the maximum line length between the left and right most 
text segments.

\subsubsection{Task Formulation}
\label{sec:col_seg_task}

It is useful to think of a table as a \textit{sequence of sequences}. In other words,
each table $t$ consists of a sequence of $n$ rows, $r_0,\hdots, r_i, \hdots, r_n$,
with each row consisting of a sequence of text and spatial features,
$r_i = \left[r_{i0} \hdots r_{ij} \hdots r_{im} \right]$.
Each row can be of varying length.
We can predict column segmentations as a
binary output for a single segment as follows: \textbf{1} if its the end of a segment, \textbf{0} otherwise.
Hence, the task is reduced to a simple binary classification problem.
\begin{equation}
  \centering
  \small
  \begin{array}{cccccc}
    Less & imputed & interest & {} & (174,862.64) & {} \\
    0 & 0 & \textbf{1} & {} & \textbf{1} & {} \\
    Less & imputed & interest &  \mathbf{\langle} \mathbf{SEG} \mathbf{\rangle}  & (174,862.64) & \mathbf{\langle} \mathbf{SEG} \mathbf{\rangle} \\
  \end{array}
\end{equation}

\subsection{Model Arsenal}



\subsubsection{Method 0: Unsupervised}
\label{sec:unsup}
Our baseline comparison will be no model, and forcing our alignment algorithm
to find disjoint columns purely on word segments in an unsupervised setting.

\subsubsection{Method 1: Feedforward}
\label{sec:feedforward}
To test if a recurrent model is necessary, we experimented with a five layer
deep feedforward model that makes predictions on a given feature set. 
We tested the effectiveness of token tag embeddings versus spatial features.
For combining the two, we use a linear layer to project the 4 spatial features to 
an intermediary embedding of size 16, concatenated to the token tag embedding, 
and feed the joint representation into the 
downstream feedforward model. 
Each layer has a hidden size of 32, with a ReLU activation. 

  

\begin{table*}
  \caption{Performance metrics per model. Considering each table as an instance, error is measured as number of detected cells versus actual number of cells. The percentiles indicate the \textit{SMAPE} for this measure. \textit{MCC} indicates the correlation between no. of identified and actual cells. \textit{\% Perfect} indicates the percentage of tables that were perfectly rendered (no errors).} \label{table:models}
  \vspace{-3mm}
  \begin{center}
  \resizebox{1.80\columnwidth}{!}{
  \begin{tabular}{ l || c || c c c c c | c | c || c || l }
    \toprule
    \textbf{Model}                         & \textbf{No. Params}   & \textbf{10\%}            & \textbf{25\%}             & \textbf{50\%}             & \textbf{75\%}             & \textbf{90\%}              & \textbf{Max}                & \textbf{\% Perfect} & \textbf{MCC}     & \textbf{Section} \\
    \midrule
    Unsupervised                  & 0            &  66.67          &  91.00           &  116.65          &  137.50          &  155.56           &  196.74            & 0.46       &  -      & \S~\ref{sec:unsup} \\
    \midrule
    Feedforward: Spatial          &  2,769       &  8.00           &  31.58           &  66.67           &  90.91           &  116.67           &  196.14            & 6.88       &  0.906  & \S~\ref{sec:feedforward}   \\    
    Feedforward: Token            & 16,801       &  8.00           &  27.79           &  50.00           &  80.00           &  110.41           &  196.61            & 6.48       &  0.623  & \S~\ref{sec:feedforward}   \\
    Feedforward: Token + Spatial  & 17,393       &  3.77           &  22.81           &  47.97           &  75.86           &  109.05           &  195.83            & 8.85       &  0.947  & \S~\ref{sec:feedforward}   \\
    \midrule
    LSTM: Row-wise                & 27,025       &  0.77           &  20.98           &  43.48           &  70.27           &  \textbf{100.00}  &  194.24            & 10.01      &  0.953  & \S~\ref{sec:row_lstm}   \\
    LSTM: Local Table             & 27,025       &  \textbf{0.00}  &  19.29           &  \textbf{41.28}  &  \textbf{69.29}  &  \textbf{100.00}  &  195.05            & 10.50      &  0.951  & \S~\ref{sec:local_lstm}   \\
    LSTM: Swapped Local Table     & 27,025       &  \textbf{0.00}  &  20.41           &  42.52           &  70.27           &  \textbf{100.00}  &  193.90            & 10.17      &  \textbf{0.954}  & \S~\ref{sec:swap_lstm} \\
    LSTM: Global Table            & 27,025       &  \textbf{0.00}  &  20.00           &  42.90           &  69.65           &  \textbf{100.00}  &  \textbf{193.53}   & 10.30      &  0.953  & \S~\ref{sec:global_lstm}   \\
    \midrule
    Transformer: Row-wise         & 27,153       &  \textbf{0.00}  &  \textbf{18.18}  &  41.88           &  69.37           &  \textbf{100.00}  &  194.24            & \textbf{10.57}      &  0.950  & \S~\ref{sec:row_trans}  \\
    Transformer: Global Table     & 27,153       &  2.35           &  22.22           &  46.15           &  73.68           &  105.20           &  195.05            &  9.71      &  0.949  & \S~\ref{sec:global_trans}  \\
    Transformer: Recurrent        & 35,761       &  \textbf{0.00}  &  22.22           &  44.44           &  71.40           &  101.66           &  194.54            & 10.04      &  0.951  & \S~\ref{sec:recurrent}  \\
    \bottomrule
  \end{tabular}
  }
  \end{center}
  \vspace{-3mm}
\end{table*}

\begin{figure*}
  \centering
  \vspace{-3mm}
    \begin{minipage}{.5\textwidth}
      \centering
      \captionof{figure}{LSTM: Row-wise}
      \label{fig:indep}
      \resizebox{!}{2.45cm}{ \begin{tikzpicture}[shorten >=1pt,node distance=1cm,on grid,auto, every node/.style={scale=0.75},
  el/.style={inner sep=2pt, align=left, sloped}]

  \node[] (r0) {$r_{i-1,0} \; \cdots \; r_{i-1,j} \; \cdots \; r_{i-1,m}$};
  \node[] (for0) [below=of r0, yshift=-0.2cm, xshift=0.75cm] {$\texttt{Forward}$};
  \node[] (rev0) [below=of r0, yshift=-1.05cm, xshift=-0.75cm] {$\texttt{Reverse}$};

  \node[] (r1) [right=of r0, xshift=3.5cm] {$r_{i,0} \; \cdots \; r_{i,j} \; \cdots \; r_{i,m}$};
  \node[] (for1) [below=of r1, yshift=-0.2cm, xshift=0.75cm] {$\texttt{Forward}$};
  \node[] (rev1) [below=of r1, yshift=-1.05cm, xshift=-0.75cm] {$\texttt{Reverse}$};

  \node[] (rn) [right=of r1, xshift=3.5cm] {$r_{i+1,0} \; \cdots \; r_{i+1,j} \; \cdots \; r_{i+1,m}$};
  \node[] (forn) [below=of rn, yshift=-0.2cm, xshift=0.75cm] {$\texttt{Forward}$};
  \node[] (revn) [below=of rn, yshift=-1.05cm, xshift=-0.75cm] {$\texttt{Reverse}$};

  \node[] (linear0) [below=of r0, yshift=-2.4cm] {$\texttt{Linear}$};
  \node[] (linear1) [below=of r1, yshift=-2.4cm] {$\texttt{Linear}$};
  \node[] (linearn) [below=of rn, yshift=-2.4cm] {$\texttt{Linear}$};

  \node[] (out0) [below=of linear0, yshift=0.15cm] {$\begin{bmatrix} 0 & 0 & \cdots & 0 \end{bmatrix}$};
  \node[] (out1) [below=of linear1, yshift=0.15cm] {$\begin{bmatrix} 0 & 1 & \cdots & 1 \end{bmatrix}$};
  \node[] (outn) [below=of linearn, yshift=0.15cm] {$\begin{bmatrix} 0 & 1 & \cdots & 0 \end{bmatrix}$};

  \path[->]
    (r0) edge [bend left=10] node[] {} (for0)
    (r0) edge [bend right=10] node[] {} (rev0)
    (for0) edge [bend left=10] node[] {} (linear0)
    (rev0) edge [bend right=10] node[] {} (linear0)
    (linear0) edge [] node[] {} (out0)

    (r1) edge [bend left=10] node[] {} (for1)
    (r1) edge [bend right=10] node[] {} (rev1)
    (for1) edge [bend left=10] node[] {} (linear1)
    (rev1) edge [bend right=10] node[] {} (linear1)
    (linear1) edge [] node[] {} (out1)

    (rn) edge [bend left=10] node[] {} (forn)
    (rn) edge [bend right=10] node[] {} (revn)
    (forn) edge [bend left=10] node[] {} (linearn)
    (revn) edge [bend right=10] node[] {} (linearn)
    (linearn) edge [] node[] {} (outn);

    \draw[red,dotted,line width=0.5mm] ($(for0.north east)+(0.1,0.45)$)  rectangle ($(rev0.south west)-(0.1,0.25)$);
    \draw[red,dotted,line width=0.5mm] ($(for1.north east)+(0.1,0.45)$)  rectangle ($(rev1.south west)-(0.1,0.25)$);
    \draw[red,dotted,line width=0.5mm] ($(forn.north east)+(0.1,0.45)$)  rectangle ($(revn.south west)-(0.1,0.25)$);
\end{tikzpicture} }
    \end{minipage}%
    \begin{minipage}{.5\textwidth}
      \centering
      \captionof{figure}{LSTM: Local Table}
      \label{fig:joint}
      \resizebox{!}{2.45cm}{ \begin{tikzpicture}[shorten >=1pt,node distance=1cm,on grid,auto, every node/.style={scale=0.75},
    el/.style={inner sep=2pt, align=left, sloped}]

    \node[] (r0) {$r_{i-1,0} \; \cdots \; r_{i-1,j} \; \cdots \; r_{i-1,m}$};
    \node[] (for0) [below=of r0, yshift=-0.2cm, xshift=0.75cm] {$\texttt{Forward}$};
    \node[] (rev0) [below=of r0, yshift=-1.05cm, xshift=-0.75cm] {$\texttt{Reverse}$};

    \node[] (r1) [right=of r0, xshift=3.5cm] {$r_{i,0} \; \cdots \; r_{i,j} \; \cdots \; r_{i,m}$};
    \node[] (for1) [below=of r1, yshift=-0.2cm, xshift=0.75cm] {$\texttt{Forward}$};
    \node[] (rev1) [below=of r1, yshift=-1.05cm, xshift=-0.75cm] {$\texttt{Reverse}$};

    \node[] (rn) [right=of r1, xshift=3.5cm] {$r_{i+1,0} \; \cdots \; r_{i+1,j} \; \cdots \; r_{i+1,m}$};
    \node[] (forn) [below=of rn, yshift=-0.2cm, xshift=0.75cm] {$\texttt{Forward}$};
    \node[] (revn) [below=of rn, yshift=-1.05cm, xshift=-0.75cm] {$\texttt{Reverse}$};

    \node[] (linear0) [below=of r0, yshift=-2.75cm] {$\texttt{Linear}$};
    \node[] (linear1) [below=of r1, yshift=-2.75cm] {$\texttt{Linear}$};
    \node[] (linearn) [below=of rn, yshift=-2.75cm] {$\texttt{Linear}$};

    \node[] (out0) [below=of linear0, yshift=0.15cm] {$\begin{bmatrix} 0 & 0 & \cdots & 0 \end{bmatrix}$};
    \node[] (out1) [below=of linear1, yshift=0.15cm] {$\begin{bmatrix} 0 & 1 & \cdots & 1 \end{bmatrix}$};
    \node[] (outn) [below=of linearn, yshift=0.15cm] {$\begin{bmatrix} 0 & 1 & \cdots & 0 \end{bmatrix}$};

    \path[->]
      (r0) edge [bend left=10] node[] {} (for0)
      (r0) edge [bend right=10] node[] {} (rev0)
      (for0) edge [bend left=10] node[] {} (linear0)
      (rev0) edge [bend right=10] node[] {} (linear0)
      (linear0) edge [] node[] {} (out0)

      (r1) edge [bend left=10] node[] {} (for1)
      (r1) edge [bend right=10] node[] {} (rev1)
      (for1) edge [bend left=10] node[] {} (linear1)
      (rev1) edge [bend right=10] node[] {} (linear1)
      (linear1) edge [] node[] {} (out1)

      (rn) edge [bend left=10] node[] {} (forn)
      (rn) edge [bend right=10] node[] {} (revn)
      (forn) edge [bend left=10] node[] {} (linearn)
      (revn) edge [bend right=10] node[] {} (linearn)
      (linearn) edge [] node[] {} (outn)

      (for0) edge [] node[] {} (for1)
      (rev0) edge [] node[] {} (rev1)
      (for1) edge [] node[] {} (forn)
      (rev1) edge [] node[] {} (revn)

      ;

      \draw[red,dotted,line width=0.5mm] ($(forn.north east)+(0.3,0.65)$)  rectangle ($(rev0.south west)-(0.3,0.45)$);
      \draw[blue,dotted,line width=0.25mm] ($(for0.north east)+(0.1,0.45)$)  rectangle ($(rev0.south west)-(0.1,0.25)$);
      \draw[blue,dotted,line width=0.25mm] ($(for1.north east)+(0.1,0.45)$)  rectangle ($(rev1.south west)-(0.1,0.25)$);
      \draw[blue,dotted,line width=0.25mm] ($(forn.north east)+(0.1,0.45)$)  rectangle ($(revn.south west)-(0.1,0.25)$);

  \end{tikzpicture} }
    \end{minipage}%
    
    \vspace{-3mm}
    \begin{minipage}{.5\textwidth}
      \centering
      \captionof{figure}{LSTM: Swapped Hidden State Propagation}
      \label{fig:swap}
      \resizebox{!}{2.45cm}{ \begin{tikzpicture}[shorten >=1pt,node distance=1cm,on grid,auto, every node/.style={scale=0.75},
  el/.style={inner sep=2pt, align=left, sloped}]

  \node[] (r0) {$r_{i-1,0} \; \cdots \; r_{i-1,j} \; \cdots \; r_{i-1,m}$};
  \node[] (for0) [below=of r0, yshift=-0.2cm, xshift=0.75cm] {$\texttt{Forward}$};
  \node[] (rev0) [below=of r0, yshift=-1.05cm, xshift=-0.75cm] {$\texttt{Reverse}$};

  \node[] (r1) [right=of r0, xshift=3.5cm] {$r_{i,0} \; \cdots \; r_{i,j} \; \cdots \; r_{i,m}$};
  \node[] (for1) [below=of r1, yshift=-0.2cm, xshift=0.75cm] {$\texttt{Forward}$};
  \node[] (rev1) [below=of r1, yshift=-1.05cm, xshift=-0.75cm] {$\texttt{Reverse}$};

  \node[] (rn) [right=of r1, xshift=3.5cm] {$r_{i+1,0} \; \cdots \; r_{i+1,j} \; \cdots \; r_{i+1,m}$};
  \node[] (forn) [below=of rn, yshift=-0.2cm, xshift=0.75cm] {$\texttt{Forward}$};
  \node[] (revn) [below=of rn, yshift=-1.05cm, xshift=-0.75cm] {$\texttt{Reverse}$};

  \node[] (linear0) [below=of r0, yshift=-2.75cm] {$\texttt{Linear}$};
  \node[] (linear1) [below=of r1, yshift=-2.75cm] {$\texttt{Linear}$};
  \node[] (linearn) [below=of rn, yshift=-2.75cm] {$\texttt{Linear}$};

  \node[] (out0) [below=of linear0, yshift=0.15cm] {$\begin{bmatrix} 0 & 0 & \cdots & 0 \end{bmatrix}$};
  \node[] (out1) [below=of linear1, yshift=0.15cm] {$\begin{bmatrix} 0 & 1 & \cdots & 1 \end{bmatrix}$};
  \node[] (outn) [below=of linearn, yshift=0.15cm] {$\begin{bmatrix} 0 & 1 & \cdots & 0 \end{bmatrix}$};

  \path[->]
    (r0) edge [bend left=10] node[] {} (for0)
    (r0) edge [bend right=10] node[] {} (rev0)
    (for0) edge [bend left=10] node[] {} (linear0)
    (rev0) edge [bend right=10] node[] {} (linear0)
    (linear0) edge [] node[] {} (out0)

    (r1) edge [bend left=10] node[] {} (for1)
    (r1) edge [bend right=10] node[] {} (rev1)
    (for1) edge [bend left=10] node[] {} (linear1)
    (rev1) edge [bend right=10] node[] {} (linear1)
    (linear1) edge [] node[] {} (out1)

    (rn) edge [bend left=10] node[] {} (forn)
    (rn) edge [bend right=10] node[] {} (revn)
    (forn) edge [bend left=10] node[] {} (linearn)
    (revn) edge [bend right=10] node[] {} (linearn)
    (linearn) edge [] node[] {} (outn)

    (for0) edge [bend left=5] node[] {} (rev1)
    (rev0) edge [bend left=5] node[] {} (for1)
    (for1) edge [bend left=5] node[] {} (revn)
    (rev1) edge [bend left=5] node[] {} (forn)
    ;

  \draw[red,dotted,line width=0.5mm] ($(forn.north east)+(0.3,0.65)$)  rectangle ($(rev0.south west)-(0.3,0.45)$);
  \draw[blue,dotted,line width=0.25mm] ($(for0.north east)+(0.1,0.45)$)  rectangle ($(rev0.south west)-(0.1,0.25)$);
  \draw[blue,dotted,line width=0.25mm] ($(for1.north east)+(0.1,0.45)$)  rectangle ($(rev1.south west)-(0.1,0.25)$);
  \draw[blue,dotted,line width=0.25mm] ($(forn.north east)+(0.1,0.45)$)  rectangle ($(revn.south west)-(0.1,0.25)$);

\end{tikzpicture} }
    \end{minipage}%
    \begin{minipage}{.5\textwidth}
      \centering
      \captionof{figure}{LSTM: Global Table}
      \label{fig:global}
      \resizebox{!}{2.45cm}{ \begin{tikzpicture}[shorten >=1pt,node distance=1cm,on grid,auto, every node/.style={scale=0.75},
    el/.style={inner sep=2pt, align=left, sloped}]

    \node[] (r0) {$r_{i-1,0} \; \cdots \; r_{i-1,j} \; \cdots \; r_{i-1,m}$};
    \node[] (for0) [below=of r0, yshift=-0.2cm, xshift=0.75cm] {$\texttt{Forward}$};
    \node[] (rev0) [below=of r0, yshift=-1.05cm, xshift=-0.75cm] {$\texttt{Reverse}$};

    \node[] (r1) [right=of r0, xshift=3.5cm] {$r_{i,0} \; \cdots \; r_{i,j} \; \cdots \; r_{i,m}$};
    \node[] (for1) [below=of r1, yshift=-0.2cm, xshift=0.75cm] {$\texttt{Forward}$};
    \node[] (rev1) [below=of r1, yshift=-1.05cm, xshift=-0.75cm] {$\texttt{Reverse}$};

    \node[] (rn) [right=of r1, xshift=3.5cm] {$r_{i+1,0} \; \cdots \; r_{i+1,j} \; \cdots \; r_{i+1,m}$};
    \node[] (forn) [below=of rn, yshift=-0.2cm, xshift=0.75cm] {$\texttt{Forward}$};
    \node[] (revn) [below=of rn, yshift=-1.05cm, xshift=-0.75cm] {$\texttt{Reverse}$};

    \node[] (linear0) [below=of r0, yshift=-2.75cm] {$\texttt{Linear}$};
    \node[] (linear1) [below=of r1, yshift=-2.75cm] {$\texttt{Linear}$};
    \node[] (linearn) [below=of rn, yshift=-2.75cm] {$\texttt{Linear}$};

    \node[] (out0) [below=of linear0, yshift=0.15cm] {$\begin{bmatrix} 0 & 0 & \cdots & 0 \end{bmatrix}$};
    \node[] (out1) [below=of linear1, yshift=0.15cm] {$\begin{bmatrix} 0 & 1 & \cdots & 1 \end{bmatrix}$};
    \node[] (outn) [below=of linearn, yshift=0.15cm] {$\begin{bmatrix} 0 & 1 & \cdots & 0 \end{bmatrix}$};

    \path[->]
      (r0) edge [bend left=10] node[] {} (for0)
      (r0) edge [bend right=10] node[] {} (rev0)
      (for0) edge [bend left=10] node[] {} (linear0)
      (rev0) edge [bend right=10] node[] {} (linear0)
      (linear0) edge [] node[] {} (out0)

      (r1) edge [bend left=10] node[] {} (for1)
      (r1) edge [bend right=10] node[] {} (rev1)
      (for1) edge [bend left=10] node[] {} (linear1)
      (rev1) edge [bend right=10] node[] {} (linear1)
      (linear1) edge [] node[] {} (out1)

      (rn) edge [bend left=10] node[] {} (forn)
      (rn) edge [bend right=10] node[] {} (revn)
      (forn) edge [bend left=10] node[] {} (linearn)
      (revn) edge [bend right=10] node[] {} (linearn)
      (linearn) edge [] node[] {} (outn)

      (for0) edge [] node[] {} (for1)
      (rev1) edge [] node[] {} (rev0)
      (for1) edge [] node[] {} (forn)
      (revn) edge [] node[] {} (rev1)

      ;

      \draw[red,dotted,line width=0.5mm] ($(forn.north east)+(0.3,0.65)$)  rectangle ($(rev0.south west)-(0.3,0.45)$);
      \draw[blue,dotted,line width=0.25mm] ($(for0.north east)+(0.1,0.45)$)  rectangle ($(rev0.south west)-(0.1,0.25)$);
      \draw[blue,dotted,line width=0.25mm] ($(for1.north east)+(0.1,0.45)$)  rectangle ($(rev1.south west)-(0.1,0.25)$);
      \draw[blue,dotted,line width=0.25mm] ($(forn.north east)+(0.1,0.45)$)  rectangle ($(revn.south west)-(0.1,0.25)$);

  \end{tikzpicture} }
    \end{minipage}%
    
    \vspace{-3mm}
    \begin{minipage}{.5\textwidth}
      \centering
      \captionof{figure}{Transformer: Row-wise}
      \label{fig:trans}
      \resizebox{!}{3.58cm}{ \begin{tikzpicture}[shorten >=1pt,node distance=1cm,on grid,auto, every node/.style={scale=0.75},
  el/.style={inner sep=2pt, align=left, sloped}, every text node part/.style={align=center}]

  \node[] (r0) {$r_{i-1,0} \; \cdots \; r_{i-1,j} \; \cdots \; r_{i-1,m}$};
  \node[] (mha0) [below=of r0, yshift=-0.4cm] {$\texttt{Multi-Head}$ \\ $\texttt{Attention}$};
  \node[] (add00) [below=of mha0, yshift=0.4cm] {$\texttt{Add \& Norm}$};
  \node[] (feed0) [below=of add00, yshift=0.3cm] {$\texttt{Feed Forward}$};
  \node[] (add01) [below=of feed0, yshift=0.63cm] {$\texttt{Add \& Norm}$};

  \node[] (r1) [right=of r0, xshift=3.5cm] {$r_{i,0} \; \cdots \; r_{i,j} \; \cdots \; r_{i,m}$};
  \node[] (mha1) [below=of r1, yshift=-0.4cm] {$\texttt{Multi-Head}$ \\ $\texttt{Attention}$};
  \node[] (add10) [below=of mha1, yshift=0.4cm] {$\texttt{Add \& Norm}$};
  \node[] (feed1) [below=of add10, yshift=0.3cm] {$\texttt{Feed Forward}$};
  \node[] (add11) [below=of feed1, yshift=0.63cm] {$\texttt{Add \& Norm}$};

  \node[] (rn) [right=of r1, xshift=3.5cm] {$r_{i+1,0} \; \cdots \; r_{i+1,j} \; \cdots \; r_{i+1,m}$};
  \node[] (mhan) [below=of rn, yshift=-0.4cm] {$\texttt{Multi-Head}$ \\ $\texttt{Attention}$};
  \node[] (addn0) [below=of mhan, yshift=0.4cm] {$\texttt{Add \& Norm}$};
  \node[] (feedn) [below=of addn0, yshift=0.3cm] {$\texttt{Feed Forward}$};
  \node[] (addn1) [below=of feedn, yshift=0.63cm] {$\texttt{Add \& Norm}$};

  \node[] (linear0) [below=of add01, yshift=0.1cm] {$\texttt{Linear}$};
  \node[] (linear1) [below=of add11, yshift=0.1cm] {$\texttt{Linear}$};
  \node[] (linearn) [below=of addn1, yshift=0.1cm] {$\texttt{Linear}$};

  \node[] (out0) [below=of linear0, yshift=0.15cm] {$\begin{bmatrix} 0 & 0 & \cdots & 0 \end{bmatrix}$};
  \node[] (out1) [below=of linear1, yshift=0.15cm] {$\begin{bmatrix} 0 & 1 & \cdots & 1 \end{bmatrix}$};
  \node[] (outn) [below=of linearn, yshift=0.15cm] {$\begin{bmatrix} 0 & 1 & \cdots & 0 \end{bmatrix}$};

  \path[->]
    (r0) edge [] node[] {} (mha0)
    (r0) edge [bend right=50] node[] {} (add00.west)
    (r0) edge [bend left=15] node[] {} ($(mha0.north east)-(0.3,0)$)
    (r0) edge [bend right=15] node[] {} ($(mha0.north west)+(0.3,0)$)
    (mha0) edge [-] node[] {} (add00)
    (add00) edge [] node[] {} (feed0)
    (feed0) edge [-] node[] {} (add01)
    (add00.east) edge [bend left=55] node[] {} (add01.east)
    (add01) edge [] node[] {} (linear0)
    (linear0) edge [] node[] {} (out0)

    (r1) edge [] node[] {} (mha1)
    (r1) edge [bend right=50] node[] {} (add10.west)
    (r1) edge [bend left=15] node[] {} ($(mha1.north east)-(0.3,0)$)
    (r1) edge [bend right=15] node[] {} ($(mha1.north west)+(0.3,0)$)
    (mha1) edge [-] node[] {} (add10)
    (add10) edge [] node[] {} (feed1)
    (feed1) edge [-] node[] {} (add11)
    (add10.east) edge [bend left=55] node[] {} (add11.east)
    (add11) edge [] node[] {} (linear1)
    (linear1) edge [] node[] {} (out1)

    (rn) edge [] node[] {} (mhan)
    (rn) edge [bend right=50] node[] {} (addn0.west)
    (rn) edge [bend left=15] node[] {} ($(mhan.north east)-(0.3,0)$)
    (rn) edge [bend right=15] node[] {} ($(mhan.north west)+(0.3,0)$)
    (mhan) edge [-] node[] {} (addn0)
    (addn0) edge [] node[] {} (feedn)
    (feedn) edge [-] node[] {} (addn1)
    (addn0.east) edge [bend left=55] node[] {} (addn1.east)
    (addn1) edge [] node[] {} (linearn)
    (linearn) edge [] node[] {} (outn);

  \draw[red,dotted,line width=0.5mm] ($(mha0.north east)+(0.5,0.5)$)  rectangle ($(add01.south west)-(0.5,0.17)$);
  \draw[red,dotted,line width=0.5mm] ($(mha1.north east)+(0.5,0.5)$)  rectangle ($(add11.south west)-(0.5,0.17)$);
  \draw[red,dotted,line width=0.5mm] ($(mhan.north east)+(0.5,0.5)$)  rectangle ($(addn1.south west)-(0.5,0.17)$);

\end{tikzpicture} }
    \end{minipage}%
    \begin{minipage}{.5\textwidth}
      \centering
      \captionof{figure}{Transformer: Recurrent}
      \label{fig:rtrans}
      \resizebox{!}{3.58cm}{ \begin{tikzpicture}[shorten >=1pt,node distance=1cm,on grid,auto, every node/.style={scale=0.75},
  el/.style={inner sep=2pt, align=left, sloped}, every text node part/.style={align=center}]

  \node[] (r0) {$r_{i-1,0} \; \cdots \; r_{i-1,j} \; \cdots \; r_{i-1,m}$};
  \node[] (mha0a) [below=of r0, yshift=-0.4cm] {$\texttt{Multi-Head}$ \\ $\texttt{Attention}$};
  \node[] (add0a) [below=of mha0a, yshift=0.4cm] {$\texttt{Add \& Norm}$};
  \node[] (mha0b) [below=of add0a, yshift=-0.2cm] {$\texttt{Multi-Head}$ \\ $\texttt{Attention}$};
  \node[] (add0b) [below=of mha0b, yshift=0.4cm] {$\texttt{Add \& Norm}$};
  \node[] (feed0) [below=of add0b, yshift=0.3cm] {$\texttt{Feed Forward}$};
  \node[] (add0c) [below=of feed0, yshift=0.63cm] {$\texttt{Add \& Norm}$};

  \node[] (r1) [right=of r0, xshift=3.5cm] {$r_{i,0} \; \cdots \; r_{i,j} \; \cdots \; r_{i,m}$};
  \node[] (mha1a) [below=of r1, yshift=-0.4cm] {$\texttt{Multi-Head}$ \\ $\texttt{Attention}$};
  \node[] (add1a) [below=of mha1a, yshift=0.4cm] {$\texttt{Add \& Norm}$};
  \node[] (mha1b) [below=of add1a, yshift=-0.2cm] {$\texttt{Multi-Head}$ \\ $\texttt{Attention}$};
  \node[] (add1b) [below=of mha1b, yshift=0.4cm] {$\texttt{Add \& Norm}$};
  \node[] (feed1) [below=of add1b, yshift=0.3cm] {$\texttt{Feed Forward}$};
  \node[] (add1c) [below=of feed1, yshift=0.63cm] {$\texttt{Add \& Norm}$};

  \node[] (rn) [right=of r1, xshift=3.5cm] {$r_{i+1,0} \; \cdots \; r_{i+1,j} \; \cdots \; r_{i+1,m}$};
  \node[] (mhana) [below=of rn, yshift=-0.4cm] {$\texttt{Multi-Head}$ \\ $\texttt{Attention}$};
  \node[] (addna) [below=of mhana, yshift=0.4cm] {$\texttt{Add \& Norm}$};
  \node[] (mhanb) [below=of addna, yshift=-0.2cm] {$\texttt{Multi-Head}$ \\ $\texttt{Attention}$};
  \node[] (addnb) [below=of mhanb, yshift=0.4cm] {$\texttt{Add \& Norm}$};
  \node[] (feedn) [below=of addnb, yshift=0.3cm] {$\texttt{Feed Forward}$};
  \node[] (addnc) [below=of feedn, yshift=0.63cm] {$\texttt{Add \& Norm}$};

  \node[] (linear0) [below=of add0c, yshift=-0.45cm] {$\texttt{Linear}$};
  \node[] (linear1) [below=of add1c, yshift=-0.45cm] {$\texttt{Linear}$};
  \node[] (linearn) [below=of addnc, yshift=-0.45cm] {$\texttt{Linear}$};

  \node[] (out0) [below=of linear0, yshift=0.15cm] {$\begin{bmatrix} 0 & 0 & \cdots & 0 \end{bmatrix}$};
  \node[] (out1) [below=of linear1, yshift=0.15cm] {$\begin{bmatrix} 0 & 1 & \cdots & 1 \end{bmatrix}$};
  \node[] (outn) [below=of linearn, yshift=0.15cm] {$\begin{bmatrix} 0 & 1 & \cdots & 0 \end{bmatrix}$};

  \path[->]
    (r0) edge [] node[] {} (mha0a)
    (r0) edge [bend right=65] node[] {} (add0a.west)
    (r0) edge [bend left=15] node[] {} ($(mha0a.north east)-(0.3,0)$)
    (r0) edge [bend right=15] node[] {} ($(mha0a.north west)+(0.3,0)$)
    (mha0a) edge [-] node[] {} (add0a)
    (add0a) edge [bend left=15] node[] {} ($(mha0b.north east)-(0.3,0)$)
    (mha0b) edge [-] node[] {} (add0b)
    (add0b) edge [] node[] {} (feed0)
    (feed0) edge [-] node[] {} (add0c)
    (add0a.east) edge [bend left=65] node[] {} (add0b.east)
    (add0b.east) edge [bend left=65] node[] {} (add0c.east)
    (add0c) edge [] node[] {} (linear0)
    (linear0) edge [] node[] {} (out0)

    (r1) edge [] node[] {} (mha1a)
    (r1) edge [bend right=65] node[] {} (add1a.west)
    (r1) edge [bend left=15] node[] {} ($(mha1a.north east)-(0.3,0)$)
    (r1) edge [bend right=15] node[] {} ($(mha1a.north west)+(0.3,0)$)
    (mha1a) edge [-] node[] {} (add1a)
    (add1a) edge [bend left=15] node[] {} ($(mha1b.north east)-(0.3,0)$)
    (mha1b) edge [-] node[] {} (add1b)
    (add1b) edge [] node[] {} (feed1)
    (feed1) edge [-] node[] {} (add1c)
    (add1a.east) edge [bend left=65] node[] {} (add1b.east)
    (add1b.east) edge [bend left=65] node[] {} (add1c.east)
    (add1c) edge [] node[] {} (linear1)
    (linear1) edge [] node[] {} (out1)

    (rn) edge [] node[] {} (mhana)
    (rn) edge [bend right=65] node[] {} (addna.west)
    (rn) edge [bend left=15] node[] {} ($(mhana.north east)-(0.3,0)$)
    (rn) edge [bend right=15] node[] {} ($(mhana.north west)+(0.3,0)$)
    (mhana) edge [-] node[] {} (addna)
    (addna) edge [bend left=15] node[] {} ($(mhanb.north east)-(0.3,0)$)
    (mhanb) edge [-] node[] {} (addnb)
    (addnb) edge [] node[] {} (feedn)
    (feedn) edge [-] node[] {} (addnc)
    (addna.east) edge [bend left=65] node[] {} (addnb.east)
    (addnb.east) edge [bend left=65] node[] {} (addnc.east)
    (addnc) edge [] node[] {} (linearn)
    (linearn) edge [] node[] {} (outn)
  ;

  \draw ($(mha0b.north)+(-2.2,0.65)$) to[out=0, in=180] ($(mha0b.north)+(-1.2,0.65)$);
  \draw [->] ($(mha0b.north)+(-1.24,0.65)$) to[out=0, in=105] ($(mha0b.north west)+(0.3,0)$);
  \draw [->] ($(mha0b.north)+(-1.24,0.65)$) to[out=0, in=105] ($(mha0b.north)$);

  \draw ($(add1c.south)$) to[out=-90, in=180] ($(add1c.south east)+(0,-0.3)$) to[out=0, in=-90] ($(mhanb.north west)+(-1.0,-1.7)$) to[out=90, in=180] ($(mhanb.north)+(-1.2,0.65)$);
  \draw [->] ($(mhanb.north)+(-1.24,0.65)$) to[out=0, in=105] ($(mhanb.north west)+(0.3,0)$);
  \draw [->] ($(mhanb.north)+(-1.24,0.65)$) to[out=0, in=105] ($(mhanb.north)$);

  \draw ($(add0c.south)$) to[out=-90, in=180] ($(add0c.south east)+(0,-0.3)$) to[out=0, in=-90] ($(mha1b.north west)+(-1.0,-1.7)$) to[out=90, in=180] ($(mha1b.north)+(-1.2,0.65)$);
  \draw [->] ($(mha1b.north)+(-1.24,0.65)$) to[out=0, in=105] ($(mha1b.north west)+(0.3,0)$);
  \draw [->] ($(mha1b.north)+(-1.24,0.65)$) to[out=0, in=105] ($(mha1b.north)$);

  \draw ($(addnc.south)$) to[out=-90, in=180] ($(addnc.south east)+(0,-0.3)$) to[out=0, in=-90] ($(addna.south east)+(0.75,-2)$);
  \draw [->] ($(addna.south east)+(0.75,-2.04)$) to[out=90, in=180] ($(addna.south east)+(1.65,-0.4)$);
  \draw [->] ($(addna.south east)+(0.75,-2.04)$) to[out=90, in=180] ($(addna.south east)+(1.65,-0.15)$);

  \draw[blue,dotted,line width=0.25mm] ($(mha0a.north east)+(0.6,0.45)$)  rectangle ($(add0c.south west)-(0.6,0.45)$);
  \draw[blue,dotted,line width=0.25mm] ($(mha1a.north east)+(0.6,0.45)$)  rectangle ($(add1c.south west)-(0.6,0.45)$);
  \draw[blue,dotted,line width=0.25mm] ($(mhana.north east)+(0.6,0.45)$)  rectangle ($(addnc.south west)-(0.6,0.45)$);

  \draw[red,dotted,line width=0.5mm] ($(mhana.north east)+(0.95,0.6)$)  rectangle ($(add0c.south west)-(0.95,0.6)$);

\end{tikzpicture} }
    \end{minipage}%
    \vspace{-4mm}
\end{figure*}

\subsubsection{Method 2: LSTM: Row-wise}
\label{sec:row_lstm}
A baseline approach is to assume independence amongst the rows.
The model is simple: embed the tokens and spatial features, 
feed the sequence into a 2-layer bidirectional
LSTM, and use a linear layer to make segment predictions.
With bidirectional layers, 
we give each token access to neighboring signals to
predict whether a new column starts or not.

\subsubsection{Method 3: LSTM: Local Table}
\label{sec:local_lstm}
Tables are sequences of sequences,
and to assume independence on individual sequences is hearsay.
When segmenting a table, readers do not focus on a
single row but makes use of the context surrounding it. Hence, a simple
yet effective improvement to induce joint row modeling is to allow for hidden
state propagation between the rows, inspired by NMT models~\cite{cho,seq2seq}. 
This \textit{deep transition} between
rows is done by passing the bidirectional
hidden states onto the next sequence, using it as the initial hidden vector~\cite{deep_trans}.
Hence, the next row will be able to process the context not only within itself,
but build upon what has already been previously seen above it,
but be able to attend \textit{locally}.

\subsubsection{Method 4: LSTM: Swapped Hidden State Propagation}
\label{sec:swap_lstm}
Another experimental model can be developed by swapping the forward and
reverse hidden states between rows. The assumption behind this model is
that recurrent networks are biased towards the most recently seen input.
Hence, during the forward computation in the next row, our hidden states
encode the context from the end of the previous row. By swapping the hidden
states, we ensure that the forward LSTM will receive the context from the
beginning of the previous row, which matches the same column pattern.

\subsubsection{Method 5: LSTM: Global Table}
\label{sec:global_lstm}
Unlike the previous cases, where we only operate a bidirectional on the row before 
propagating the hidden state information to the next, the global model processes 
the entire table as a singular sequence. So, for a segment in the first row, 
it incorporates information from the tokens before it in the same row, and 
all further table tokens downstream in the reverse direction. Hence, it has a 
\textit{global} scope over the previous \textit{localized} models.

\subsubsection{Method 6: Transformer: Row-Wise}
\label{sec:row_trans}
In recent years, Vaswani et al.~\cite{attention_is_all_you_need} introduced a
completely new paradigm for sequence modeling, the \textit{transformer}. Instead
of iterative computation and propagating hidden states, a \textit{transformer}
uses a self-attention mechanism for a key, value, and query.
For our purposes, we incorporate the encoder architecture that can
operate on a single row and produce independent column alignments, for
a direct comparison to the row-wise LSTM model.

\subsubsection{Method 7: Transformer: Global Table}
\label{sec:global_trans}
We also feed the entire table as a singular sequence. Hence, the multi-head attention 
can look throughout different rows and jointly make predictions for probable column 
segmentations. This is considered the \textit{global} approach to joint table modeling.

\subsubsection{Method 8: Transformer: Recurrent}
\label{sec:recurrent}
However, we again encounter the same issue of a \textit{global} over \textit{local}
approach to joint modeling.
Hence, we experiment with a version of the transformer decoder that can be 
both recurrent and attend locally for each row. 
In the original decoder, a target embedding sequence would mix with the output
of the encoder for a given source sentence. However, using the traditional
encoder-decoder architecture would only provide pairwise row training. 
To achieve a joint model, we 
refeed the decoder's output in place of the original encoder's input. Hence,
the decoder can first attend \textit{locally} to the row in the 
first multi-head attention layer,
then use the \textit{memory} of the previous step as the key and value.
To initialize the \textit{memory} state, we allow for a 
globally learned vector that
can be expanded to the correct sequence length.

\subsubsection{Experiment Details}
We used a 50/50 train-validation split, 
resulting in two equal sets of 1,519 tables. Models were trained using the 
logistic loss on the segments directly. 
We used the Adam optimizer~\cite{kingma2014adam} with $\alpha = 0.01$ and 
trained for 640 parameter updates, saving the model with the best validation  
matthews correlation coefficient (MCC)~\cite{MATTHEWS1975442}. 
This metric is a balanced measure given our 
class imbalance was 44\% positive, 56\% negative.
Interestingly, both sets achieved similar MCC scores 
indicating that our training
set accurately represents the average use case.

\subsubsection{Pipeline Incorporation}
In order to integrate our model into the alignment pipeline,
we first convert the text from the Tesseract metadata into
the 881 valid tokens.
Next, we calculate the spatial information for each segment. We query our 
trained model with these features, and receive a prediction per segment.
Using these, we concatenate segments that 
are predicted to be merged together.

\subsection{Alignment Evaluation \& Discussion}
Several trends are apparent from the results presented in Table~\ref{table:models}. 
First, sequential modeling out performs both feedforward models and the unsupervised approach. 
Second, LSTM's benefit from seeing the full table, unlike transformers that do better in a \textit{local} scope. 
Transformer-based models may provide too much access to each segment during the MHA,
unlike LSTM's that bias towards a segments local neighborhood. 
Finally, \textit{local} joint table modeling outperforms \textit{global} modeling, as 
our models can attend to the content that is most relevant, without additional noise from a larger scope.
Therefore, the best model for column segmentation, in conjunction with out alignment algorithm, is the 
\textbf{LSTM: Local Table}.


\section{Conclusion}
This paper has outlined an end-to-end pipeline to \textit{discover}, \textit{extract}, and
\textit{format} tabular content from \textit{image} documents, with high fidelity.
Leveraging advanced deep learning techniques across several sub-fields has enabled us to effectively
solve each sub-task, culminating in a strong pipeline that provides a valuable
transcription service for tabular data.
We foresee this pipeline as integral to many downstream applications.
For instance, table values can now be indexed by aligning the row field and
column header to a particular value. Rendered excel sheets can be quickly corrected
by users, and directly embedded to financial projections and equations.

\bibliographystyle{ACM-Reference-Format}
\bibliography{sample-base}

\end{document}